\documentclass[final,3p,times,twocolumn]{elsarticle}

\usepackage{amssymb}
\usepackage{amsmath}
\usepackage{multirow}

\journal{Expert Systems with Applications}
\usepackage{booktabs}
\usepackage{longtable}
\usepackage{amsmath}
\usepackage{xcolor}
\usepackage{algorithm}  
\usepackage{algorithmic}  
\usepackage{pifont}
\usepackage{hyperref}
\usepackage[normalem]{ulem}

\newcommand{\add}[1]{{\color{blue}{#1}}}
\renewcommand{\add}[1]{#1}

\definecolor{top1}{RGB}{255,0,0} 
\definecolor{top2}{RGB}{0,0,255} 
\definecolor{top3}{RGB}{0,255,0} 

\begin{document}

\begin{frontmatter}



\title{FlexiD-Fuse: Flexible number of inputs multi-modal medical\\
image fusion based on diffusion model}




\author[author1]{Yushen Xu}
\ead{2112355010@stu.fosu.edu.cn}

\author[author1,label1,label2]{Xiaosong Li\corref{cor}}
\ead{lixiaosong@buaa.edu.cn}

\author[author1]{Yuchun Wang}
\ead{2112455008@stu.fosu.edu.cn}

\author[label2]{Xiaoqi Cheng}
\ead{chengxiaoqi@fosu.edu.cn}

\author[author2]{Huafeng Li}
\ead{lhfchina99@kust.edu.cn}

\author[author1]{Haishu Tan}
\ead{tanhaishu@fosu.edu.cn}


\cortext[cor]{Corresponding author}


\address[author1]{ School of Physics and Optoelectronic Engineering, Foshan University, Foshan 528225, China}

\address[label1]{Guangdong-HongKong-Macao Joint Laboratory for Intelligent Micro-Nano Optoelectronic Technology, Foshan 528225, China}

\address[label2]{Guangdong Provincial Key Laboratory of Industrial Intelligent Inspection Technology, Foshan University, Foshan 528000, China}

\address[author2]{ School of  Information Engineering and Automation, Kunming University of Science and Technology, Kunming 650500, China}

\begin{abstract}
Different modalities of medical images provide unique physiological and anatomical information for diseases. Multi-modal medical image fusion integrates useful information from different complementary medical images with different modalities, producing a fused image that comprehensively and objectively reflects lesion characteristics to assist doctors in clinical diagnosis. However, existing fusion methods can only handle a fixed number of modality inputs, such as accepting only two-modal or tri-modal inputs, and cannot directly process varying input quantities, which hinders their application in clinical settings. To tackle this issue, we introduce FlexiD-Fuse, a diffusion-based image fusion network designed to accommodate flexible quantities of input modalities. It can end-to-end process two-modal and \add{tri-modal} medical image fusion under the same weight. FlexiD-Fuse transforms the diffusion fusion problem, which supports only fixed-condition inputs, into a maximum likelihood estimation problem based on the diffusion process and hierarchical Bayesian modeling. By incorporating the Expectation-Maximization algorithm into the diffusion sampling iteration process, FlexiD-Fuse can generate high-quality fused images with cross-modal information from source images, independently of the number of input images. We compared the latest two-modal and tri-modal medical image fusion methods, tested them on Harvard datasets, and evaluated them using nine popular metrics. The experimental results show that our method achieves the best performance in medical image fusion with varying inputs. Meanwhile, we conducted extensive extension experiments on infrared and visible image fusion, multi-exposure image fusion, and multi-focus image fusion tasks with arbitrary numbers, and compared them with the perspective state-of-the-art (SOTA) methods. The results of the extension experiments consistently demonstrate the effectiveness and superiority of our method. The code is available at \href{https://github.com/XylonXu01/FlexiD-Fuse}{https://github.com/XylonXu01/FlexiD-Fuse}.
\end{abstract}




\end{frontmatter}



\section{Introduction}
With the development of medical imaging technology, common imaging modalities include computed tomography (CT)\cite{1}, magnetic resonance imaging (MRI) \cite{jie2023medical, 3}, positron emission tomography (PET) \cite{4}, and single-photon emission computed tomography (SPECT) \cite{5}. These imaging modalities each provide different anatomical, functional, or metabolic information \cite{1R1}. However, due to differences in imaging principles and objectives, single-modal images often fail to capture the complex characteristics of biological tissues comprehensively \cite{6,1R3}. For example, CT images can clearly show structural details of bones and high-density tissues but perform poorly in soft tissue imaging \cite{7}. MRI excels in soft tissue contrast and spatial resolution, but cannot accurately display bones or calcified tissues \cite{8}. Functional imaging modalities such as PET and SPECT can provide real-time data on metabolic activity and physiological functions within the body \cite{1R4,1R5}, but they have low spatial resolution and are inadequate for depicting anatomical details \cite{9, 10}. These differences mean that relying on single-modal images for diagnosis or treatment planning often leads to incomplete information, thereby impacting diagnostic accuracy and therapeutic effectiveness \cite{11, 12}. Multi-modal medical image fusion technology integrates the advantages of different medical imaging modalities to generate a fused image that simultaneously presents anatomical, functional, and metabolic information, aiding clinical diagnosis and treatment decision making \cite{13, 14}. For example, fusing MRI and CT images can simultaneously reveal details of both soft and hard tissues \cite{15}. Li \cite{16} designed a dual-path convolutional segmentation model to achieve the  fusion segmentation of any two different modal images. Guo et al. \cite{17} developed a convolutional neural network that combines CT, MRI, and PET image data for contour segmentation of soft-tissue sarcomas. \add{Karthik et al. \cite{karthik2024ensemble} proposed combining  MRI tomography and PET to provide multi-modal images for image segmentation of soft tissue tumor lesions. }\add{The advancement of flexible fusion methods, as proposed in this paper, further enables clinicians to select and fuse the most relevant imaging modalities according to specific patient needs, thus optimizing personalized diagnosis and treatment strategies.}

Currently, multi-modal image fusion models are mainly divided into two categories: two-modal image fusion and tri-modal image fusion. For two-modal image fusion models, main approaches include fusion methods based on autoencoders \cite{18,19,liu2024infrared,li2021different}, convolutional neural networks (CNNs) \cite{li2025mulfs,22,23,24,25}, transformer models \cite{26,27,li2025fefdm,29,30}, generative adversarial networks (GANs) \cite{31,32, 1R2}, and Diffusion models \cite{33,34,35}. Among these, GAN-based fusion models often exhibit excellent performance but suffer from instability and mode collapse during training, which can result in decreased fusion image quality and unreasonable distributions \cite{33}. In contrast, Diffusion-based methods effectively address these instability and mode collapse issues. However, due to the lack of ground truth, traditional Diffusion models cannot be directly applied to image fusion \cite{36}. Moreover, existing diffusion models are characterized by high computational costs and prolonged fusion times, which pose significant challenges in fulfilling the stringent requirements of clinical applications. In tri-modal image fusion models, the primary fusion methods are divided into traditional methods \cite{9,37} and deep learning-based methods \cite{35,38}. Traditional methods include sparse representation-based approaches \cite{9} and multi-scale transformation-based approaches \cite{37}. These methods typically require designing complex fusion rules for targeted feature extraction, often lacking robustness when handling two-modal fusion. Deep learning methods include Diffusion-based approaches \cite{35} and GAN-based approaches \cite{38}.

In summary, current medical image fusion methods typically support a fixed number of modalities, such as two or three. However, in clinical practice, physicians need the flexibility to select the number of modalities based on the specific needs of the patient. When only two modalities are required for diagnosis, using three modalities not only increases the time and financial burden on patients but also introduces unnecessary complexity. Conversely, when three modalities are required for diagnosis (as shown in Figure \ref{Fig1}.(a)), an alternate fusion strategy is employed, wherein two source images are fused first, followed by progressive fusion with the remaining image until all images are combined. Although this strategy is feasible, multiple rounds of feature extraction and fusion may lead to information loss, resulting in a final fused image that deviates from reality and increases the risk of misdiagnosis. Furthermore, existing deep learning-based fusion methods often demand high computational resources, while traditional methods rely on complex feature extraction designs. When adapting two-modal methods to tri-modal scenarios or vice versa, these approaches often experience performance degradation and increased time costs. 


\begin{figure}[htbp]
\centering	
\includegraphics[width=1.0\linewidth]{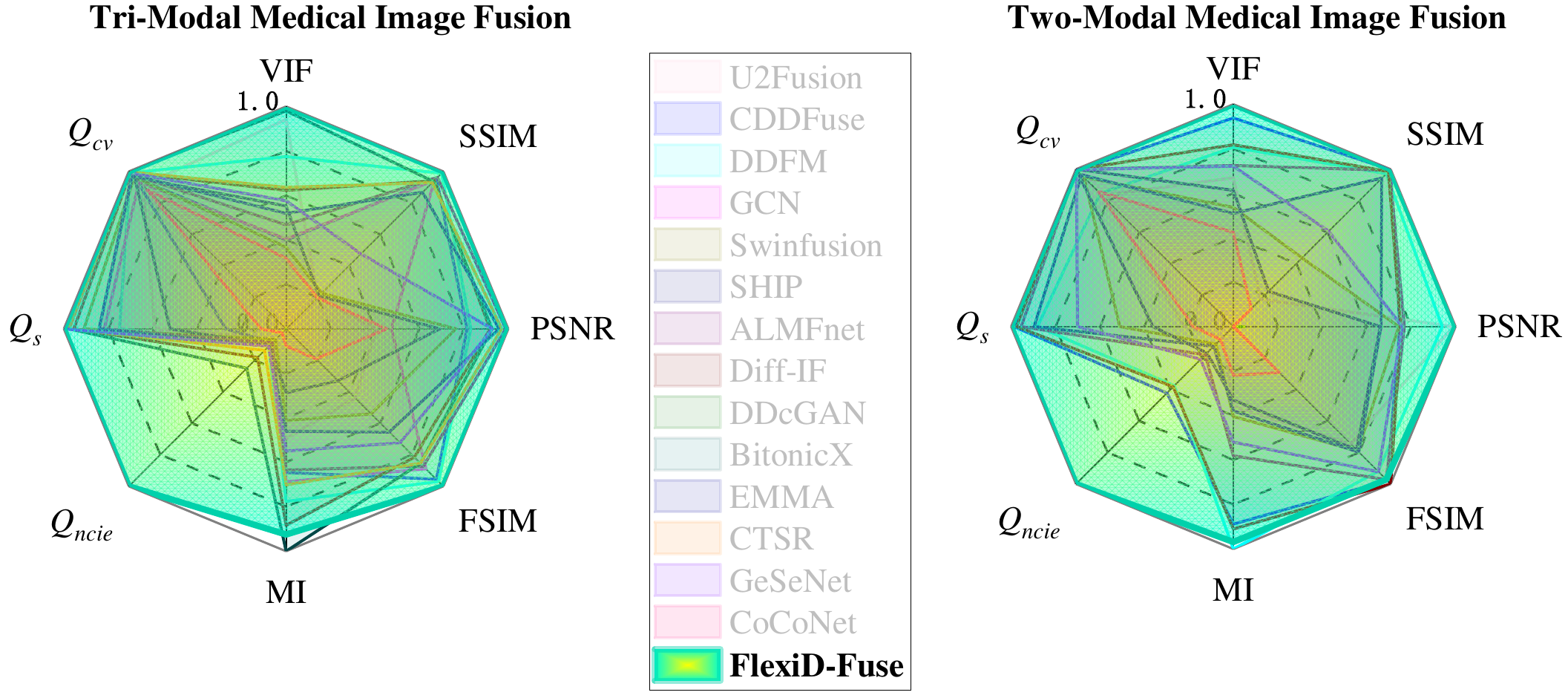}
\caption{\add{A comparison of results with the most advanced methods in the fusion of two-modal and three-modal medical images. The radar map highlights the superiority of the FlexiD-Fuse.}}
\label{Fig.Radar}
\end{figure}

Therefore, we propose FlexiD-Fuse, a diffusion process-based arbitrary-modal medical image fusion model. As shown in Figure \ref{Fig1}(b), this method transforms the diffusion fusion problem, which only supports fixed condition inputs, into a training-free maximum likelihood estimation problem based on diffusion processes and hierarchical Bayesian modeling. Each modal image undergoes separate Bayesian modeling, and the expectation-maximization algorithm is integrated into the diffusion sampling iteration process. Meanwhile, \add{we introduce} an ultra-lightweight diffusion fusion mamba for prior modeling of source images, effectively \add{increasing} the inference speed. \add{Figure \ref{Fig.Radar} indicates that the proposed FlexiD-Fuse outperforms fifteen SOTA methods in both three-modal medical image fusion and two-modal medical image fusion}. Our contributions can be summarized as follows:
\begin{enumerate}
    \item \add{We propose FlexiD-Fuse, an end-to-end diffusion model-based medical image fusion framework. By incorporating hierarchical Bayesian modeling and expectation maximization, our method flexibly accommodates varying input modalities and yields high-quality fused images under unified weights, overcoming the fixed-input limitations in existing methods.}
    \item \add{We propose an ultra-lightweight diffusion fusion Mamba, significantly accelerates the inference speed of diffusion models during the fusion progress.}
    \item \add{In two-modal and tri-modal medical image fusion tasks, our method outperforms the SOTA fusion methods in both visual quality and quantitative evaluations.}
    \item \add{We further extend FlexiD-Fuse to infrared and visible, multi-exposure, and multi-focus image fusion tasks and conduct comparative experiments with 16 methods across six datasets. The results consistently demonstrate the robustness and superiority of our method.}
\end{enumerate}

\section{Related Work}
In this section, we will introduce existing methods for multi-modal image fusion, as well as methods capable of fusing an arbitrary number of modalities.
\subsection{Diffusion Models}
Diffusion models have occupied a leading position in the field of visual generation by virtue of their excellent performance\cite{63,controlNet}. \add{These models} constructs a Markov chain by progressively adding noise in the forward process to estimate the latent data distribution and generate images through the reverse sampling process. The inherent denoising property of this model enables diffusion models to exhibit outstanding and stable performance in visual restoration tasks\cite{zhang2024text,yang2025lfdt,36}. Nevertheless, diffusion models encounter two major challenges in image fusion applications: one is the requirement for high-quality ground truth images, and the other is the relatively slow generation speed. Recent research has focused on how to combine diffusion models with unsupervised image fusion and explore integration methods of different model architectures with diffusion models.

For example, in the aspect of unsupervised image fusion, DDFM \cite{ddfm} formulates the fusion task as a conditional generation problem based on the DDPM sampling framework, which is further subdivided into an unconditional generation sub-problem and a maximum likelihood sub-problem. VDMUFusion \cite{VDMUFusion}, on the contrary, formulates image fusion as a weighted averaging process and integrates the fusion problem into the diffusion sampling process by establishing appropriate assumptions for the noise in the diffusion model. In terms of the combination of model architectures, DiT \cite{DiT} replaces the traditional U-Net backbone with a Transformer operating on latent patches, reducing the complexity of the diffusion model. ZigMa \cite{ZigMa} surpasses Transformer-based diffusion models in terms of speed and memory utilization by introducing the Mamba spatial state model. Our method, based on ZigMa, adopts a unidirectional scanning strategy to further enhance generation speed. 

\subsection{Two-modal Fusion Methods}
In the past, two-modal image fusion was the most popular approach, and researchers proposed numerous two-modal fusion algorithms. These algorithms can be broadly categorized into traditional methods and deep learning-based methods. Among traditional methods, those based on multi-scale transforms \cite{39,40,41} and sparse representation \cite{42,43,44} are representative. Multi-scale transform methods, such as pyramid transforms \cite{45,46} and wavelet transforms \cite{47}, decompose source images from multiple perspectives to obtain multi-scale information, fuse the sub-images, and then perform inverse transformations to generate the fusion result. Sparse representation methods construct a complete dictionary and use dictionary elements to replace the source images for fusion. However, these methods have two major drawbacks: first, manually designed fusion strategies have limited capability to preserve the complementary information of source images; second, the same transformation is applied to inputs of different modalities, ignoring their specific characteristics.

In recent years, deep learning-based image fusion methods have gradually become mainstream. In autoencoder (AE)-based fusion algorithms, Zhao et al. \cite{20} proposed CDDFuse, which adopts a branched Transformer-CNN architecture as the encoder and decoder to share and extract modal features. In CNN-based image fusion algorithms, Xu et al. \cite{48} proposed U2Fusion, which employs a CNN architecture as a densely connected network and utilizes weight blocks to obtain data-driven weights for retaining features from different source images. In Transformer-based fusion algorithms, Ma et al. \cite{26} proposed SwinFusion, a universal image fusion framework based on cross-domain remote learning and the Swin Transformer. 

\begin{figure*}[htb]
\centering	
\includegraphics[width=1.0\linewidth]{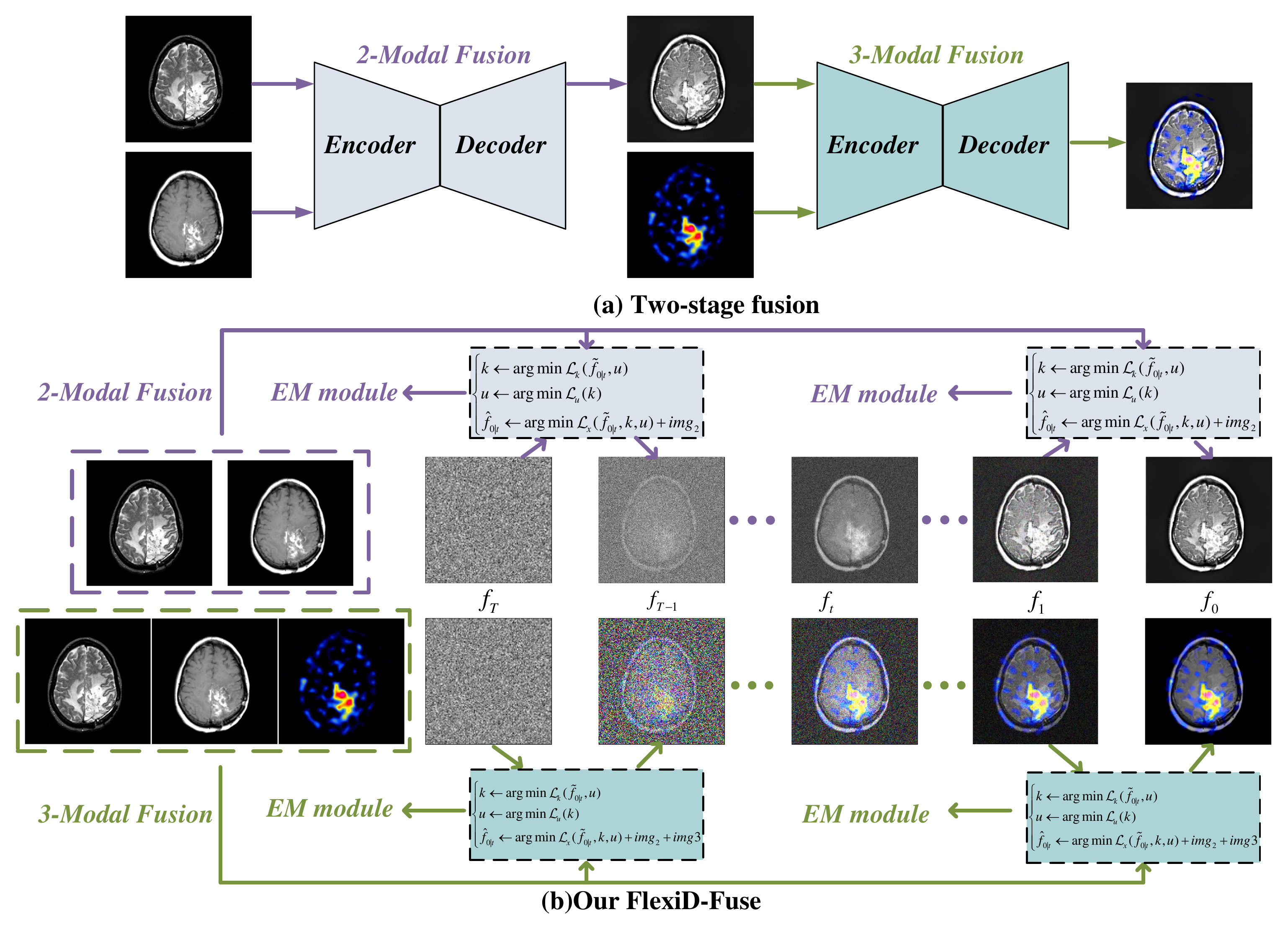}
\caption{Fusion strategies for different numbers of modalities: (a) The existing two-stage fusion process; (b) The proposed FlexiD-Fuse fusion process: first, fused features \(f_t\) are generated through the DFM model, and then the likelihood correction algorithm is applied to adjust the DFM output using information from the source images.}
\label{Fig1}
\end{figure*}

\subsection{Tri-Modal Fusion Methods}
Compared to two-modal fusion, tri-modal medical image fusion holds greater significance in clinical practice \cite{9,17,34,36,37}. For instance, Chen et al. \cite{49} developed a three-pathway network architecture that utilizes SPECT, CT, and contrast-enhanced CT (CCT) images for parathyroid hyperplasia detection. However, relatively few methods exist for tri-modal medical image fusion. Among traditional approaches, Jie et al. \cite{9} proposed a tri-modal medical image fusion method based on sparse representation and adaptive energy selection. This method fuses cartoon components using an adaptive energy selection scheme and merges texture components through sparse representation. Subsequently, they \cite{37} proposed a tri-modal medical image fusion and denoising method based on BitonicX filtering, which achieves both image fusion and denoising by analyzing energy, pixel gradients, and sharpness. Among deep learning-based methods, Huang et al. \cite{38} proposed an end-to-end generative adversarial network for tri-modal medical image fusion, leveraging multi-scale compression and an excitation-inference attention network to enhance the fusion of tri-modal images. Xu et al. \cite{35} introduced a conditional diffusion probability-based model for simultaneous tri-modal fusion and super-resolution, providing an end-to-end solution for both tasks.

\subsection{Free Quantity Fusion Methods}
Currently, methods that support image fusion with an arbitrary number of inputs are relatively rare, with GRFusion, proposed by Li et al.\cite{50} being one of them. GRFusion is a multi-focus image fusion method that allows independent detection of focal attributes for each source image, enabling simultaneous fusion of multiple source images and avoiding information loss caused by alternating fusion. However, since GRFusion requires fusion via an input multi-focus decision map, it cannot effectively address multi-modal image fusion tasks.

Existing image fusion methods often struggle with fusion tasks involving an arbitrary number of modalities, and mainstream approaches require different models for each modality, which is both cumbersome and suboptimal. Therefore, it is necessary to introduce an innovative diffusion process-based framework for arbitrary modal medical image fusion, enabling simpler applications in clinical practice.

\section{Method}
\begin{figure*}[htbp]
\centering	
\includegraphics[width=1.0\linewidth]{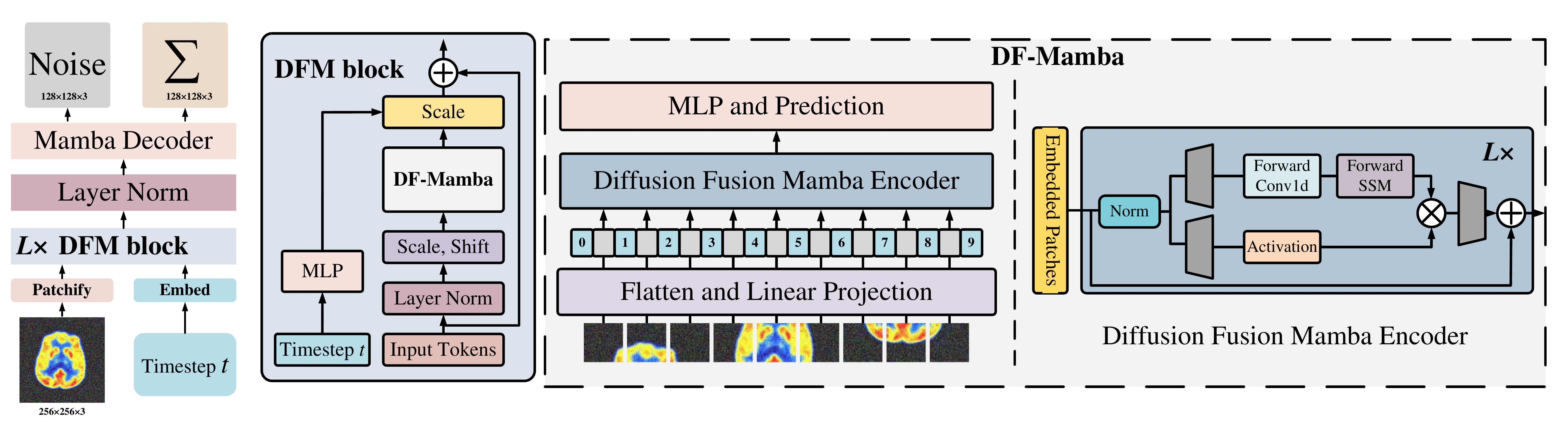}
\caption{The DFM architecture is illustrated. First, the input image is divided into multiple patches and projected into patch tokens. Subsequently, the generated token sequence is fed into the proposed DFM encoder. Unlike the bidirectional modeling in Vim, the DFM encoder processes the token sequence in a unidirectional manner to ensure faster output.}
\label{Fig2}
\end{figure*}
This section introduces the workflow of FlexiD-Fuse (as shown in Figure \ref{Fig1} (b)). We elaborate on the fusion of \add{flexible numbers of modalities} from the perspectives of the image diffusion process and expectation-maximization algorithm, while introducing the ultra-lightweight Diffusion Fusion Mamba (DFM) model.
\subsection{Problem Formulation}
Previous medical image fusion methods typically define the fusion task as taking two or more source images (e.g. \(I_{MRI},I_{CT}\) or \(I_{MRI-T1}, I_{MRI-T2},I_{SPECT}\)) as input and using a fusion network \(\theta_n\) to generate a fused image that combines relevant information from all modalities. The network is designed to learn a predefined fusion function \(\mathrm{F}_{mf} \). Mathematically, this can be expressed as:
\begin{equation}
I_{f_2}=\mathrm{F}_{mf_2} (I_{MRI},I_{CT};\theta_{n_2} )
\end{equation}
\begin{equation}
    I_{f_3}=\mathrm{F}_{mf_3} (I_{MRI-T1}, I_{MRI-T2},I_{SPECT};\theta_{n_3} )
\end{equation}
where \(I_{f_2}\) and \(I_{f_3}\) denote the two-modal and three-modal fused images. \(\theta_{n_2}\) and \(\theta_{n_3}\) represent the fusion networks for two-modal and three-modal inputs. \(\mathrm{F}_{mf_2}\) and \(\mathrm{F}_{mf_3}\) are the predefined fusion functions for two-modal and three-modal scenarios.

This formula indicates that the fusion network is optimized to learn a fixed fusion strategy tailored to specific modalities. However, in real-world medical applications, the number of modalities may vary, making this task paradigm inflexible and suboptimal in such scenarios.

To address these issues, we explore a more general approach capable of handling an arbitrary number of modalities and adapting to various image qualities. The new formula can be described as:
\begin{equation}
    I_f=\mathrm{F}_{gf} (I_{mod_1},\ldots,I_{mod_n};\theta_g )
\end{equation}
here \(\mathrm{F}_{gf}\) represents a general fusion function capable of fusing an arbitrary number of input modalities \(I_{mod_1},\ldots,I_{mod_n}\), where \(n=2\) or \(n=3\), \(\theta_g\) represents the generalized fusion network. Our approach allows the fusion network to dynamically adjust its fusion strategy based on the available modalities, providing a more flexible and robust solution for fusion tasks.
\subsection{Score-based Diffusion Process}

This paper employs a score-based diffusion process for image fusion. During pretraining, clean samples\(f_0\) are progressively perturbed with noise to approach a Gaussian signal \(f_T\). This forward process can be described by an Ito stochastic differential equation \cite{51}:
\begin{equation}
    df = -\frac{\alpha(t)}{2}x_t \, dt + \sqrt{\alpha(t)} \, dw
\end{equation}
where \(dw\) represents standard Brownian motion, and \(\alpha(t)\) is the predefined noise schedule of the Stochastic Differential Equation (SDE).

The forward process can be time-reversed and retains the form of an SDE \cite{anderson1982reverse}:
\begin{align}
\mathrm{d}\boldsymbol{x} = \left[-\frac{\alpha(t)}{2}\boldsymbol{x}_t - \alpha(t)\nabla_{\boldsymbol{x}_t} \log p_t(\boldsymbol{x}_{\boldsymbol{t}})\right] \mathrm{d}t + \sqrt{\alpha(t)}\mathrm{d}\overline{\boldsymbol{w}},
\label{eq4}
\end{align}
the unknown par \(\nabla_{\boldsymbol{x}_t} \log p_t(\boldsymbol{x}_t)\), can be modeled using the so-called \textit{score function} \(\boldsymbol{s}_{\theta}(\boldsymbol{x}_t)\) via denoising score matching techniques. This score function can be trained by optimizing the following objective:
\begin{align}
    \mathbb{E}_t \mathbb{E}_{\boldsymbol{x}_0} \mathbb{E}_{\boldsymbol{x}_t|\boldsymbol{x}_0} \left[ \left\| \boldsymbol{s}_{\boldsymbol{\theta}}(\boldsymbol{x}_t, t) - \nabla_{\boldsymbol{x}_t} \log p_{0t}(\boldsymbol{x}_t|\boldsymbol{x}_0) \right\|_2^2 \right]
\end{align}
where \(t\) is uniformly sampled over \([0,T]\) and the data pair \((\boldsymbol{x}_0,\boldsymbol{x}_t) \sim p_0(\boldsymbol{x})p_{0t}(\boldsymbol{x}_t|\boldsymbol{x}_0)\).

For the medical image fusion task with a flexible number of modal inputs, inspired by Formula \ref{eq4}, the time-reversed SDE takes the form:
\begin{equation}
\begin{split}
df = & \left[ -\frac{\alpha(t)}{2}f 
    - \alpha(t) \nabla_f \log{p_t}\left(f_t \mid I_{mod_1}, \ldots, I_{mod_n}\right) \right] dt \\
    & + \sqrt{\alpha(t)} \, d\bar{w}
\end{split}
\end{equation}
where \(d\bar{w}\) corresponds to the backward process of standard Brownian motion, and the score function \(\nabla_f \log{p_t}\left(f_t \mid I_{mod_1}, \ldots, I_{mod_n}\right)\) can be expressed as:
\begin{align}
&\nabla_f \log{p_t}\left(f_t \mid I_{mod_1}, \ldots, I_{mod_n}\right) 
\\ &= \nabla_{f_t} \log{p_t}\left(f_t\right) 
+ \nabla_{f_t} \log{p_t}\left(I_{mod_1}, \ldots, I_{mod_n} \mid f_t\right) \\
&\approx \nabla_f \log{p_t}\left(f_t\right) 
+ \nabla_f \log{p_t}\left(I_{mod_1}, \ldots, I_{mod_n} \mid {\widetilde{f}}_{0|t}\right)
\label{eq7}
\end{align}
where \({\widetilde{f}}_{0|t}\) is the estimate of \(f_0\) in the unconditional DDPM given \(f_t\). This equation is derived from Bayes' theorem, and the approximate formulation has been validated in \cite{52}.

In Equation \ref{eq7}, the first term represents the score function for unconditional diffusion sampling, which can be derived using DFM, as detailed in Section \ref{3.3}. In Section \ref{3.4}, we provide a detailed explanation of the method to obtain \(\nabla_f \log{p_t}\left(I_{mod_1}, \ldots, I_{mod_n} \mid {\widetilde{f}}_{0|t}\right)\).

\subsection{Diffusion Fusion Mamba}
\label{3.3}


We present an ultra-lightweight architecture, designated as DFM, specifically tailored for the diffusion fusion task. This architecture is grounded on the DiT framework of AdaLN \cite{DiT}, since DiT has been attested in the literature as a scalable and efficient structure \cite{ZigMa,OpenAI2024Sora}. The design of DFM adheres strictly to the standard Mamba architecture \cite{53} to ensure the preservation of its lightweight property. In view of our concentration on diffusion probabilistic models (DDPM) in image fusion, especially the spatial representation of images, DFM adopts the Vim architecture \cite{54} and operates on patch sequences \cite{55}. Figure \ref{Fig2} exhibits the overall architecture of DFM. This section elaborates the forward propagation process of DFM and the various components within its design space.

\subsubsection{Preliminaries of Mamba}
State space models \cite{56,57,58}, such as state space sequence models (S4) \cite{59} and Mamba \cite{53}, aim to encode and decode one-dimensional sequence inputs. These models are inspired by continuous systems, encoding a one-dimensional function or sequence \(x\left(t\right)\in R\) via a hidden state \(h\left(t\right)\in R^N\), and decoding it into an output signal \(y(t)\) using ordinary differential equations (ODEs).
\begin{equation}
h'_t = \mathbf{A} h_t + \mathbf{B} x_t
\end{equation}
\begin{equation}
y_t = \mathbf{C} h_t
\end{equation}
where \(\mathbf{A}\in R^{N\times N}\) as the evolution parameter, while \(\mathbf{B}\in R^{N\times1}\) and \(\mathbf{C}\in R^{1\times N}\) function as projection parameters.

Typically, inputs in natural language and 2D vision are discrete signals, so Mamba employs the zero-order hold (ZOH) rule for discretization.
\begin{equation}
\overline{\mathbf{A}} = \exp(\Delta \mathbf{A})
\end{equation}
\begin{equation}
\overline{\mathbf{B}} = (\Delta \mathbf{A})^{-1} \left( \exp(\Delta \mathbf{A}) - \mathbf{I} \right) \cdot \Delta \mathbf{B}
\end{equation}
\begin{equation}
h_t = \overline{\mathbf{A}} h_{t-1} + \overline{\mathbf{B}} x_t
\end{equation}
\begin{equation}
y_t = \mathbf{C} h_t
\end{equation}
where \(\mathbf{A}\) and \(\mathbf{B}\) represent the discrete parameters after ZOH transformation, and \(\Delta\) denotes the step size.
Finally, the model output is computed using global convolution:
\begin{equation}
\overline{\mathbf{P}} = (\mathbf{C}\overline{\mathbf{B}}, \mathbf{C}\overline{\mathbf{A}\mathbf{B}}, \ldots, \mathbf{C}\overline{\mathbf{A}}^{M-1}\overline{\mathbf{B}}), \quad y = \mathbf{x} * \overline{\mathbf{P}}
\end{equation}
where \(M\) is the length of the input sequence \(\mathbf{x}\), and \(\overline{\mathbf{P}}\) belongs to the structured convolution kernel.
\subsubsection{Patchify}
The input to DFM is a \(256 \times 256 \times 3\) tensor. Since the Mamba architecture is primarily designed for processing one-dimensional inputs, it faces challenges in learning the two-dimensional data structure of images. To address this, the first layer of DFM is "Patchify," which performs a linear embedding on each patch of the input to convert the spatial input into a token sequence of \(T\) tokens, each with a dimension of \(d\). After patchification, sinusoidal-cosine positional embeddings \cite{55} are applied to all input tokens. The number of tokens \(T\) generated during patchification is determined by the hyperparameter \(p\), which defines the patch size.
\subsubsection{Diffusion Fusion Mamba (DFM) block}
After the patchify operation, the input tokens are processed through a series of DFM blocks. In addition to handling noisy image inputs, the fusion diffusion model also needs to process additional conditional information, such as the noise timestep \(t\). Therefore, we made slight but significant modifications to the standard Vim block design. The design of all blocks is illustrated in Figure \ref{Fig2}.

Unlike Vim, DF-Mamba adopts unidirectional sequence modeling to make the model more lightweight and improve processing speed. The timestep \(t\) and the token sequence \(x_{l-1}\) are simultaneously input into the DFM block. The timestep \(t\) undergoes two layers of MLP for dimensional alignment and feature transformation, resulting in \(t'\). The token sequence \(x_{l-1}\) is first normalized through a normalization layer to obtain \(x'_{l-1}\), and then \(x'_{l-1}\) is projected onto \(z\) and \(y\), each with a dimension of \(E\).

Next, a 1-D convolution is \add{applied to} \(z\) to obtain \(z'_i\), which is then linearly projected to \(\mathbf{B}_i\), \(\mathbf{C}_i\) and \(\Delta_i\), respectively. In the SSM, \(z_{forward}\) is computed using \(\mathbf{A}_i\) transformed by \(\Delta_i\) and \(\mathbf{B}_i\). Then, \(z_{forward}\) is gated through \(y\), and they are summed to produce the output sequence \(x'_l\). Finally, the SSM output sequence \(x'_l\) is added to the timestep \(t'\) to produce \(x_l\).

\subsubsection{Mamba Decoder}
After the final DFM block, we decode the SSM output sequence into the predicted noise output and the predicted diagonal covariance. The shapes of the two decoded outputs match the original spatial input. To perform this decoding, we employ a standard linear decoder. Specifically, we apply AdaLN normalization, then linearly decode each token into a tensor of shape \(p \times p \times 2c\), where \(c\) represents the channel count of the spatial input to the DFM. Finally, the decoded tokens are rearranged back into their original spatial layout to produce the predicted noise and covariance.
\subsection{Expectation-maximization(EM) module}
\label{3.4}
\begin{figure}[htbp]
\centering	
\includegraphics[width=1.0\linewidth]{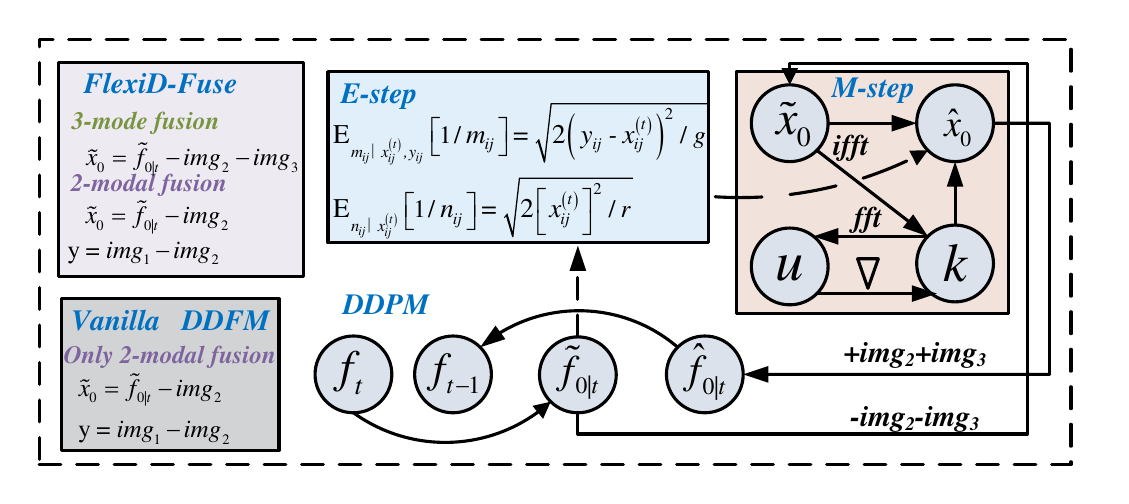}
\caption{Flowchart of FlexiD-Fuse computed in one iteration.}
\label{Fig3}
\end{figure}
The key to FlexiD-Fuse's capability of handling arbitrary modalities lies in the EM module, which updates \({\widetilde{f}}_{0|t}\) to \({\hat{f}}_{0|t}\). DDFM demonstrated the feasibility of the EM module in Diffusion, and this paper extends the EM module to support medical image fusion with an arbitrary number of modalities.

The proof here focuses on the seamless switching and fusion of two modalities and three modalities. First, we propose the loss function for this seamless switching fusion task:
\begin{equation}
    \min_f \left( \|f - \text{img}_1 - \text{img}_3\|_1 
+ \phi \|f - \text{img}_2 - \text{img}_3\|_1 \right)
\label{eq:15}
\end{equation}
The proposed loss function effectively addresses the optimization problem when fusing different numbers of modalities. When fusing two modalities, \(img_3=0\), i.e:
\begin{equation}
    \min_f \left( \|f - \text{img}_1\|_1 + \phi \|f - \text{img}_2\|_1 \right)
\end{equation}

Then, by performing a simple variable substitution \(x=f-img_2-img_3\) and \(y=img_1-img_2\), we derive the optimization formula:
\begin{equation}
    \min_x \left( \|y - x\|_1 + \phi \|x\|_1 \right)
    \label{17}
\end{equation}
It is evident that \(y\) is a known quantity, while \(x\) is an unknown quantity. The optimization problem in Equation (\ref{17}) can be transformed into a maximum likelihood inference problem using probabilistic analysis \cite{60}. The log-likelihood function for the probabilistic inference problem in image fusion is:
\begin{equation}
\begin{aligned}
l(x) &= \log p(x, y) - r(x) \\
&= -\sum_{i,j} \left[ \frac{\left(x_{ij} - y_{ij}\right)^2}{2m_{ij}} 
+ \frac{x_{ij}^2}{2n_{ij}} \right] 
- \frac{\psi}{2} \|\nabla x\|_2^2
\end{aligned}
\label{18}
\end{equation}
where \(r(x)=\|\nabla x\|_2^2\) denotes the total variation penalty term \cite{61,62}, and \(x\) is assumed to follow a Laplace distribution. The variables \(m_{ij}\) and \(n_{ij}\) serve as latent variables within the Bayesian framework.

To solve this maximum log-likelihood problem, as referenced in \cite{33}, it can be treated as a latent variable optimization problem, using the EM algorithm to obtain the optimal \(x\)

\subsubsection{E-step}
In the E-step, the probability distribution of each modal image is computed, i.e., the expectation \(Q\) of the log-likelihood function with respect to \(p\left(a,b\mid x^{\left(t\right)},y\right)\):
\begin{equation}
Q\left(x \mid x^{(t)}\right) = \mathbb{E}_{a,b \mid x^{(t)}, y}\left[l\left(x\right)\right]
\end{equation}
The conditional expectations of the latent variables \(1/m_{ij}\) and \(1/n_{ij}\) in Equation(\ref{18}):
\begin{equation}
\mathbb{E}_{m_{ij} \mid x_{ij}^{(t)}, y_{ij}}\left[\frac{1}{m_{ij}}\right] 
= \frac{2(y_{ij} - x_{ij}^{t})^2}{\gamma}
\label{20}
\end{equation}
\begin{equation}
\mathbb{E}_{n_{ij} \mid x_{ij}^{(t)}}\left[\frac{1}{n_{ij}}\right] 
= \sqrt{\frac{2\left[x_{ij}^{(t)}\right]^2}{\rho}}
\label{21}
\end{equation}
Based on Equations (\ref{20}) and (\ref{21}), the \(Q\) function can be derived as:
\begin{align}
Q &= -\sum_{i,j} \left[ \frac{\tilde{m}_{ij}}{2}(x_{ij} - y_{ij})^2 + \frac{\tilde{n}_{ij}}{2} x_{ij}^2 \right] - \frac{\psi}{2} \|\nabla \mathbf{x}\|_2^2 \\
&\propto -\|m \odot (x - y)\|_2^2 - \|n \odot {x}\|_2^2 - \psi \|\nabla {x}\|_2^2
\end{align}
where \(\widetilde{m}\) and \(\widetilde{n}\) represent \(E_{m_{ij}\mid x_{ij}^{\left(t\right)},y_{ij}}\left[\frac{1}{m_{ij}}\right]\) and \(E_{n_{ij}\mid x_{ij}^{\left(t\right)}}\left[\frac{1}{n_{ij}}\right]\). \(\odot\) denotes element-wise multiplication. Each element of matrices \(m\) and \(n\) is represented by \(\sqrt{{\widetilde{m}}_{ij}}\) and \(\sqrt{{\widetilde{n}}_{ij}}\), respectively.
\subsubsection{M-step}
The M-Step is used to determine the parameters of the distribution. First, we minimize the negative \(Q\) function with respect to \(x\) using the half-quadratic splitting algorithm, i.e.,:
\begin{equation}
\min_{x, u, k} \|m \odot (x - y)\|_2^2 
+ \|n \odot x\|_2^2 + \psi \|u\|_2^2,
\end{equation}
\[\text{s.t.} \quad u = \nabla k, \quad k = x.\]
then it is transformed into an unconstrained optimization problem, i.e.,:
\begin{equation} 
\begin{aligned}
\min_{x, u, k} \quad & \|m \odot (x - y)\|_2^2 
+ \|n \odot x\|_2^2 + \psi \|u\|_2^2 \\
&+ \frac{\eta}{2} \left( \|u - \nabla k\|_2^2 + \|k - x\|_2^2 \right).
\end{aligned}
\label{25}
\end{equation}

The unknown variables \(k\), \(u\), and \(x\) can be solved using Coordinate Descent:
\begin{equation}
\begin{aligned}
\left\{
\begin{aligned}
&\min_k \mathcal{L}_k = \|u - \nabla k\|_2^2 + \|k - x\|_2^2, \\
&\min_u \mathcal{L}_u = \frac{\eta}{2} \|u - \nabla k\|_2^2 + \psi \|u\|_2^2, \\
&\min_x \mathcal{L}_x = \|m \odot (x - y)\|_2^2 + \|n \odot x\|_2^2 
+ \frac{\eta}{2} \|k - x\|_2^2.
\end{aligned}
\right.
\end{aligned}
\end{equation}

The solutions for k, u, and s are:
\begin{equation}
\left\{
\begin{aligned}
k &= \text{ifft} \left\{ \frac{\text{fft}(x) + \overline{\text{fft}(\nabla)}
 \odot \text{fft}(u)}{1 + \overline{\text{fft}(\nabla)}
 \odot \text{fft}(\nabla)} \right\}, \\
u &= \frac{\eta}{2\psi + \eta} \nabla k, \\
x &= \left( 2m^2 \odot y + \eta k \right) \oslash \left( 2m^2 + 2n^2 + \eta \right)^{-1}.
\label{27}
\end{aligned}
\right.
\end{equation}
where \(\text{fft}\) denotes the Fast Fourier Transform, \(\overline{.}\) represents the complex conjugate, \(\oslash\) represents element-wise division.

The final estimation of \(f\) is:
\begin{align}
\hat{f} &= x + \text{img}_2 + \text{img}_3 \\
&\text{If 2-mode fusion, } \text{img}_3 = 0
\end{align}

Additionally, the hyper-parameters \(\gamma\) and \(\rho\) in Equations (\ref{20}) and (\ref{21}) can also be updated following the sampling from \(x\) (as shown in Equation (\ref{27})) by
\begin{equation}
    \gamma = \frac{1}{hw} \sum_{i,j} \mathbb{E}[m_{ij}], \quad \rho = \frac{1}{hw} \sum_{i,j} \mathbb{E}[n_{ij}]
\end{equation}
where \(hw\) respectively stand for the height and width of the image.
\subsection{FlexiD-Fuse framework}
In summary, we propose a loss function tailored for two-modal and tri-modal image fusion, derive its hierarchical Bayesian model, and perform inference using the EM algorithm. This section introduces the training and inference process of FlexiD-Fuse.
\subsubsection{Train}
During the training phase, we pretrain the denoising network using six modalities of medical images: MRI-T1, MRI-T2, CT, SPECT, PET, and Gad, enabling it to acquire prior knowledge of medical images. This process follows the same procedure as DDPM training \cite{63} and does not involve the EM model.
\subsubsection{Inference}
\begin{algorithm}[t]
\caption{FlexiD-Fuse}
\label{alg:DDFM}
\begin{algorithmic}[1]
\REQUIRE ~~\\
    \STATE $img_1$, $img_2$, $img_3$
\IF{Two-mode fusion}
    \STATE $img_3 = 0$
\ENDIF
\ENSURE ~~\\
Fused image $f_0$
\STATE $f_T \sim \mathcal{N}(0, I)$
\FOR{$t = T - 1, \cdots, 0$}
    \STATE \% \textbf{Obtain} $\tilde{f}_{0|t}$
    \STATE $\hat{x} \leftarrow z_\theta(f_t, t)$
    \STATE $\tilde{f}_{0|t} \leftarrow \frac{1}{\sqrt{\alpha_t}} \left(f_t + (1 - \alpha_t) \hat{x} \right)$
    \STATE \% \textbf{E-step}
    \IF{Three-mode fusion}
        \STATE $\tilde{x}_0 = \tilde{f}_{0|t} - img_2 - img_3, \; y = img_1 - img_2$
    \ELSIF{Two-mode fusion}
        \STATE $\tilde{x}_0 = \tilde{f}_{0|t} - img_2, \; y = img_1 - img_2$
    \ENDIF
    \STATE Evaluate expectations:
    
        \(E_{m_{ij}\mid x_{ij}^{\left(t\right)},y_{ij}}\left[\frac{1}{m_{ij}}\right]=\sqrt{\frac{2\left(y_{ij}-x_{ij}^{\left(t\right)}\right)^2}{\gamma}}\)

        \(E_{n_{ij}\mid x_{ij}^{\left(t\right)}}\left[\frac{1}{n_{ij}}\right]=\sqrt{\frac{2\left[x_{ij}^{\left(t\right)}\right]^2}{\rho}}\)
    \STATE \% \textbf{M-step}
    
    \(k = \text{ifft} \left\{ \frac{\text{fft}(\mathbf{x}) + \overline{\text{fft}(\nabla)} \odot \text{fft}(\mathbf{u})}{1 + \overline{\text{fft}(\nabla)} \odot \text{fft}(\nabla)} \right\}\)

    \(u = \frac{\eta}{2 \psi + \eta} \nabla \mathbf{k}\)

    \(x = \left( 2 {m}^2 \odot {y} + \eta {k} \right) \oslash \left( 2 {m}^2 + 2 {n}^2 + \eta \right)\)

    \STATE $\hat{f}_{0|t} = x + img_2 + img_3$
    \STATE \% \textbf{Estimate} $f_{t-1}$
    \STATE $z \sim \mathcal{N}(0, I)$
    \STATE $f_{t-1} \leftarrow \frac{\sqrt{\alpha_t(1 - \alpha_{t-1})}}{1 - \alpha_t} f_t + \frac{\sqrt{\alpha_{t-1}}\beta_t}{1 - \alpha_t} \hat{f}_{0|t} + \tilde{\sigma}_t z$
    
\ENDFOR
\end{algorithmic}
\end{algorithm}
During the inference phase, there are two main components: the diffusion model and the EM module. First, the pretrained diffusion model provides prior knowledge of medical images, enhancing confidence in the fused medical images. Then, the EM module automatically analyzes the number of modalities and adjusts the likelihood to refine the diffusion model's output, preserving information from different modal images. The specific process is detailed in Algorithm \ref{alg:DDFM}.

\section{Experiments}
In this section, we present detailed experiments on two-modal and tri-modal medical image fusion to demonstrate the superiority of our method in fusing images with varying numbers of modalities.
\subsection{Experimental Detail}
\subsubsection{Dataset and Metrics}
For the medical image fusion experiments, we used the Harvard Medical Image Dataset \cite{summers2003harvard}. For the two-modal fusion experiments, we used a total of 810 image pairs, with 70\% (567 pairs) allocated for training the FlexiD-Fuse model to learn the pixel distributions of medical images, and 30\% (243 pairs) used to evaluate model performance. These image pairs include MRI/CT, MRI/PET, and MRI/SPECT. For the tri-modal fusion experiments, a total of 119 image pairs were randomly divided into 84 pairs for the training set and 35 pairs for the test set. These image pairs include MR-T2/MR-Gad/PET, CT/MR-T2/SPECT, MR-T1/MR-T2/PET, MR-T2/MR-Gad/SPECT, and MR-T1/MR-T2/SPECT modalities.

We used nine metrics for the comprehensive evaluation of the fusion results, including Structural Similarity Index Measure (SSIM), Mutual Information (MI), Peak Signal-to-Noise Ratio (PSNR), Visual Information Fidelity (VIF), \(\mathrm{Q_{S}}\) \cite{65}, \(\mathrm{Q_{CV}}\) \cite{66}, FSIM \cite{67}, and \add{\(\mathrm{Q_{NCIE}}\)} \cite{68}.
\begin{table*}[ht]
\centering
\caption{Objective Metrics for Tri-modal Medical Image Fusion (\textcolor{top1}{Red}: Best, \textcolor{top2}{Blue}: Second, and \textcolor{top3}{Green}: Third)}
\begin{tabular}{llcccccccc}
\toprule
{Method} & {Source} & {VIF↑} & {SSIM↑} & {PSNR↑} & {FSIM↑} & {MI↑} & {\(\mathrm{Q_{NCIE}}\)↑} & {\(\mathrm{Q_{S}}\)↑} & {\(\mathrm{Q_{CV}}\)↓} \\
\midrule
U2Fusion   & TPAMI'22         & \textcolor{top3}{8.9095} & 0.1753 & 47.396 & 0.8721 & 0.5950 & 0.8082 & 0.8968 & 2075.899 \\
CDDFuse    & CVPR'23          & 4.0950 & 0.6221 & 48.049 & \textcolor{top3}{0.8894} & 0.6308 & 0.8085 & 0.8954 & 1821.209 \\
DDFM       & ICCV'23 (Oral)   & 7.4326 & \textcolor{top2}{0.6416} & 48.076 & \textcolor{top2}{0.8899} & 0.6617 & 0.8088 & 0.8975 & 1642.850 \\
GCN        & JBHI'24          & 4.1390 & 0.6151 & 47.505 & 0.8872 & 0.6408 & 0.8087 & 0.8978 & 1948.991 \\
SwinFusion & JAS'22           & 3.8142 & 0.1916 & 47.968 & 0.8755 & 0.5756 & 0.8083 & 0.8982 & 1790.075 \\
SHIP       & CVPR'24          & 5.4545 & 0.1811 & 47.850 & 0.8685 & 0.5458 & 0.8087 & 0.8965 & 1799.289 \\
ALMFnet    & TCSVT'23         & 4.7124 & \textcolor{top3}{0.6253} & \textcolor{top2}{48.327} & 0.8867 & 0.6282 & 0.8085 & 0.8979 & 1693.167 \\
Diff-IF    & Inform Fusion'24 & 6.0841 & 0.6130 & 48.215 & 0.8848 & \textcolor{top3}{0.6871} & \textcolor{top3}{0.8096} & 0.8984 & 1641.880 \\
DDcGAN     & TIP'20           & 0.5666 & 0.0576 & 46.746 & 0.8574 & 0.4780 & 0.8075 & 0.8942 & 7059.556 \\
BitonicX   & TIM'23           & \textcolor{top2}{9.2453} & 0.6099 & 48.228 & 0.8860 & \textcolor{top1}{0.7146} & \textcolor{top2}{0.8104} & \textcolor{top3}{0.8985} & \textcolor{top3}{1603.137} \\
EMMA       & CVPR'24          & 5.1668 & 0.5703 & 48.277 & 0.8794 & 0.5869 & 0.8088 & 0.8979 & 1714.620 \\
CTSR       & Measurement'22   & 6.2016 & 0.6056 & \textcolor{top3}{48.298} & 0.8860 & 0.6437 & 0.8091 & \textcolor{top3}{0.8985} & \textcolor{top2}{1603.006} \\
GeSeNet     & TNNLS'23    & 5.6648  & 0.3395  & 48.230  & 0.8817 & 0.6078  & 0.8088  & \textcolor{top1}{0.8986}  & 1638.575  \\
CoCoNet     & IJCV'23     & 3.3710  & 0.1752  & 47.468  & 0.8638  & 0.4961  & 0.8078  & 0.8947  & 2241.298 \\
Ours       & Proposed         & \textcolor{top1}{9.4243} & \textcolor{top1}{0.6474} & \textcolor{top1}{48.345} & \textcolor{top1}{0.8909} & \textcolor{top2}{0.6969} & \textcolor{top1}{0.8191} & \textcolor{top1}{0.8986} & \textcolor{top1}{1596.948} \\
\bottomrule
\end{tabular}
\label{tab:2}
\end{table*}
\begin{figure*}[h!]
\centering	
\includegraphics[width=1.0\linewidth]{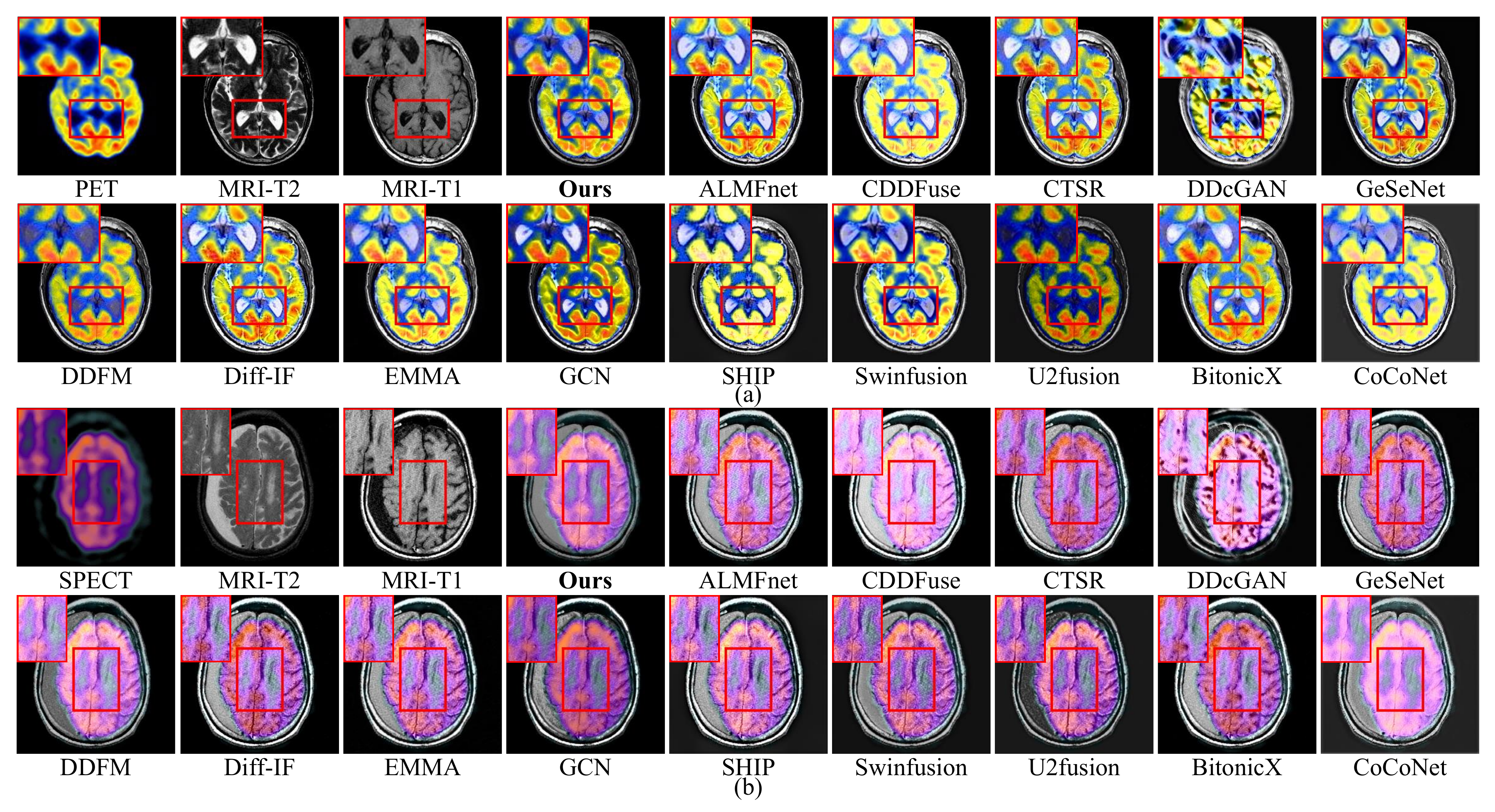}
\caption{Results of Tri-modal Medical Image Fusion.(a): PET/MRI-T2/MRI-T1 Fusion, (b): SPECT/MRI-T2/MRI-T1 Fusion}
\label{Fig4}
\end{figure*}
\begin{table*}[htbp]
\centering
\caption{Objective Metrics for Two-modal Medical Image Fusion (\textcolor{top1}{Red}: Best, \textcolor{top2}{Blue}: Second, and \textcolor{top3}{Green}: Third)}
\begin{tabular}{llcccccccc}
\toprule
{Method} & {Source} & {VIF↑} & {SSIM↑} & {PSNR↑} & {FSIM↑} & {MI↑} & {\(\mathrm{Q_{NCIE}}\)↑} & {\(\mathrm{Q_{S}}\)↑} & {\(\mathrm{Q_{CV}}\)↓} \\
\midrule
U2Fusion   & TPAMI'22         & 0.2758 & 0.2536 & \textcolor{top2}{16.777} & 0.760 & 0.636 & 0.8053 & 0.3531 & 937.048 \\
CDDFuse    & CVPR'23          & \textcolor{top2}{0.3697} & \textcolor{top2}{0.7484} & 15.366 & \textcolor{top2}{0.796} & 0.884 & \textcolor{top2}{0.8104} & \textcolor{top2}{0.8990} & 281.723 \\
DDFM       & ICCV'23 (Oral)   & 0.3236 & 0.7328 & \textcolor{top3}{16.491} & 0.787 & \textcolor{top1}{0.936} & 0.8097 & 0.8240 & 883.323 \\
SwinFusion & JAS'22           & 0.2287 & 0.4112 & 15.340 & 0.767 & 0.657 & 0.8059 & 0.5601 & 462.521 \\
SHIP       & CVPR'24          & 0.2557 & 0.2837 & 14.852 & 0.766 & 0.623 & 0.8067 & 0.4550 & \textcolor{top1}{260.863} \\
ALMFnet    & TCSVT'23         & 0.2947 & \textcolor{top3}{0.7461} & 15.483 & \textcolor{top3}{0.795} & 0.740 & 0.8069 & 0.8762 & \textcolor{top3}{262.876} \\
EMMA       & CVPR'24          & 0.2196 & 0.7166 & 14.871 & 0.768 & 0.645 & 0.8061 & 0.8351 & 404.063 \\
Diff-IF    & Inform Fusion'24 & \textcolor{top3}{0.3277} & 0.7419 & 15.390 & \textcolor{top1}{0.799} & \textcolor{top3}{0.894} & \textcolor{top3}{0.8098} & \textcolor{top3}{0.8911} & 270.354 \\
DDcGAN     & TIP'20           & 0.0414 & 0.1487 & 11.261 & 0.646 & 0.468 & 0.8043 & 0.1972 & 4344.589 \\
GeSeNet     & TNNLS'23        & 0.2937 & 0.5135 & 15.4241  & 0.7873  & 0.7113  & 0.8073  & 0.6962  & 270.098  \\
CoCoNet     & IJCV'23           & 0.1900  & 0.2165  & 10.9696  & 0.6907  & 0.5707  & 0.8055  & 0.3250  & 838.296  \\
Ours       & Proposed         & \textcolor{top1}{0.3914} & \textcolor{top1}{0.7493} & \textcolor{top1}{16.839} & 0.795 & \textcolor{top2}{0.921} & \textcolor{top1}{0.8187} & \textcolor{top1}{0.9087} & \textcolor{top2}{261.561} \\
\bottomrule
\end{tabular}
\label{tab:3}
\end{table*}
\subsubsection{Implement details}
We used an NVIDIA GeForce RTX 3090 GPU for image fusion. In Equation.\ref{25}, \(\eta\) and \(\varphi\) were set to 0.1 and 0.5, respectively. The images were converted to grayscale and normalized to the range [-1, 1] before being input into the model. The diffusion step count \(T\) was set to 100.
\subsubsection{Comparison with SOTA methods}
In this section, we compare our FlexiD-Fuse with \add{fourteen} modal medical image fusion SOTA methods, covering the three tri-modal fusion models: BitonicX \cite{37}, CTSR \cite{9}, and GCN \cite{38}, and the \add{eleven} two-modal fusion models: U2Fusion \cite{48}, CDDFuse \cite{20}, DDFM \cite{ddfm}, SwinFusion \cite{26}, SHIP \cite{69}, ALMFnet \cite{70}, EMMA \cite{22}, Diff-IF \cite{37}, DDcGAN \cite{32}, \add{GeSeNet \cite{Gesenet} and CoCoNet \cite{CoCoNet}}, For the two-modal fusion models, we adopt a two-stage fusion process to handle tri-modal medical image fusion.

\subsection{Tri-modal Medical Image Fusion}
\subsubsection{Objective Evaluation}
The comparison of objective evaluation metrics for tri-modal fusion experiments with existing SOTA methods (as shown in Table \ref{tab:2}) demonstrates that FlexiD-Fuse excels in multiple key metrics, showcasing its advantages in tri-modal medical image fusion tasks. Overall, whether compared to models specifically designed for tri-modal fusion or two-modal fusion models, FlexiD-Fuse demonstrates outstanding performance across all models, with significant advantages in detail preservation, noise suppression, information fusion, and subjective visual quality.
\subsubsection{Subjective Evaluation}

Figure \ref{Fig4} illustrates the fusion results for PET/MRI-T2/MRI-T1 and SPECT/MRI-T2/MRI-T1 images, comparing FlexiD-Fuse with various existing fusion methods.

In terms of detail preservation, whether fusing PET or SPECT with MRI, FlexiD-Fuse excels at retaining the structural details of MRI-T1 and MRI-T2. Compared with CTSR and DDcGAN, FlexiD-Fuse demonstrates higher clarity in high-frequency regions (e.g., brain boundaries and structural lines), while CTSR and DDcGAN exhibit some degree of blurring or smoothing. DDFM and Diff-IF are overly smooth in the red-boxed areas, whereas FlexiD-Fuse maintains sharp edges. Regarding functional information preservation, FlexiD-Fuse performs well in metabolic activity regions of PET and functional areas of SPECT, avoiding information loss issues seen in GCN and SwinFusion. It also effectively addresses the unnatural over-brightness in certain areas caused by excessive enhancement in ALMFnet and CTSR, ensuring balanced representation of functional information. In terms of multi-modal information balance, FlexiD-Fuse achieves balanced fusion of structural information (MRI-T1, MRI-T2) and functional information (PET or SPECT), with natural transitions, especially in boundary regions. In contrast, U2Fusion and BitonicX prioritize structural information during fusion, leading to a weakening of functional information. In terms of visual consistency and naturalness, FlexiD-Fuse exhibits smooth and natural transitions in the PET or SPECT and MRI transition areas, whereas BitonicX and CTSR show segmentation or abrupt transitions.

\subsection{Two-modal Medical Image Fusion}
In this section, we compare our FlexiD-Fuse with 9 SOTA methods for two-modal medical image fusion, including U2Fusion, CDDFuse, DDFM, SwinFusion, SHIP, ALMFnet, DDcGAN, EMMA, and Diff-IF.

\subsubsection{Objective Evaluation}

Table \ref{tab:3} presents the objective evaluation metrics for FlexiD-Fuse in two-modal medical image fusion, demonstrating its excellent performance across multiple key metrics. The VIF and PSNR results indicate that FlexiD-Fuse retains more image information during fusion, minimizes detail loss, and significantly improves the visual quality of fused images. Additionally, its outstanding performance in the SSIM metric highlights FlexiD-Fuse's ability to balance contributions from each modality, producing sharp, high-quality fused images. The \(\mathrm{Q_{CV}}\) and \(Q_s\) metrics further confirm that FlexiD-Fuse not only integrates multi-source information effectively but also preserves key image features, enhancing contrast and clarity.

\subsubsection{Subjective Evaluation}
\begin{figure*}[htbp]
\centering	
\includegraphics[width=1.0\linewidth]{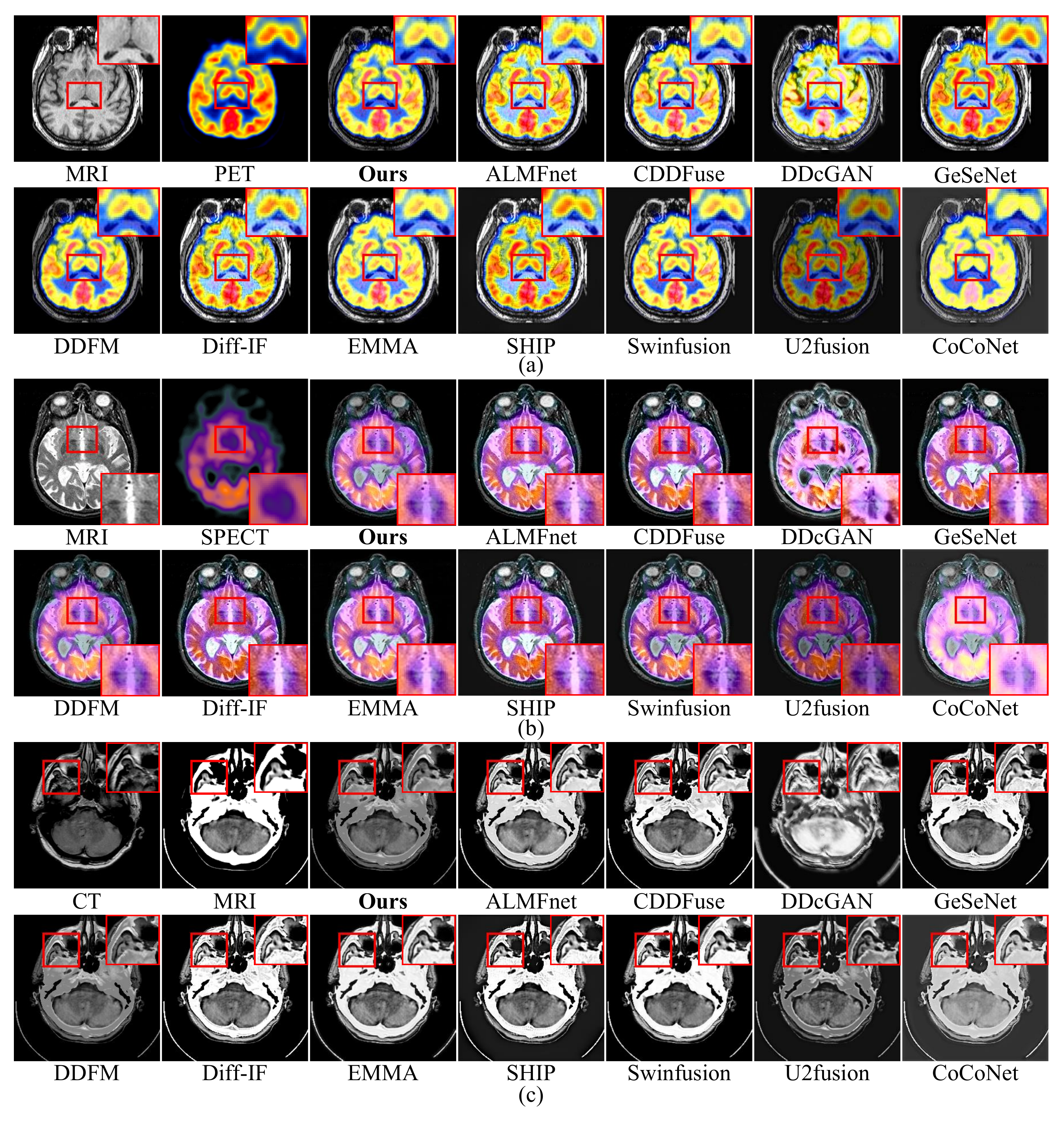}
\caption{Results of Two-modal Medical Image Fusion.(a): MRI-PET Fusion, (b): MRI-SPECT Fusion, (c):CT-MRI Fusion}
\label{Fig5}
\end{figure*}
Figure \ref{Fig5} illustrates the comparative results of different methods in two-modal medical image fusion: Figure \ref{Fig5}(a) shows MRI-PET fusion, Figure \ref{Fig5}(b) shows MRI-SPECT fusion, and Figure \ref{Fig5}(c) shows CT-MRI fusion.

\begin{table*}[h!]
\centering
\caption{\add{Comparison of Parameter Scale, Inference Time, and Performance among Multi-modal Fusion Models (2M: Two-modal Fusion; 3M: Tri-modal Fusion)} }
\label{tab:my-table}
\resizebox{\textwidth}{!}{%
\begin{tabular}{llllllll}
\hline
Method     & Parameters (M) & FLOPs-2M (G) & Time-2M (s) & SSIM-2M & FLOP-3M & Time-3M (s) & SSIM-3M \\
\hline
CDDFuse    & 1.190           & 29.212        & 0.017    & 0.748  & 58.420   & 0.034    & 0.622  \\
DDFM       & 552.810         & 1114.711      & 34.501    & 0.732   & 2229.420 & 65.002       & 0.641   \\
GCN        & 0.043          & Null         & Null     & Null    & 7.640    & 0.864    & 0.615   \\
SwinFusion & 0.970           & 62.821         & 2.772     & 0.411   & 125.610   & 5.544     & 0.191   \\
SHIP       & 0.548          & 401.51       & 5.980    & 0.283  & 803.021  & 11.965    & 0.181  \\
ALMFnet    & 0.053          & 3.590         & 0.053   & 0.746   & 7.181    & 0.106    & 0.625   \\
Diff-IF    & 23.710          & 43.530        & 0.290    & 0.741  & 87.061   & 0.581     & 0.613  \\
EMMA       & 1.520           & 8.912          & 0.064    & 0.716   & 17.821    & 0.128    & 0.570  \\
GeSeNet    & 0.241           & 16.236       & 0.0005   & 0.148  & 32.472  & 0.001    & 0.339  \\
CoCoNet    & 9.132           & 10.391        & 0.015    & 0.216  & 20.781   & 0.031     & 0.175  \\
Ours       & 174.891         & 54.192        & 9.850    & \textbf{0.749}   & 108.384 & 9.850    & \textbf{0.647}   \\ \hline
\end{tabular}%
}
\end{table*}

FlexiD-Fuse effectively preserves anatomical details and enhances functional contrast in multi-modal medical image fusion. It retains sharp structural information, such as clear tissue boundaries in MRI/CT, outperforming SwinFusion and U2Fusion, which produce blurred or oversaturated edges. Compared to DDcGAN and ALMFnet, FlexiD-Fuse exhibits superior preservation of spatial anatomical details critical for medical diagnosis. In terms of functional imaging, it successfully integrates PET metabolic activity and SPECT blood flow, ensuring clear, well-defined, color-coded intensity without the over-enhancement or under-saturation issues observed in DDFM and EMMA. FlexiD-Fuse maintains a balanced representation of intensity and detail, avoiding exaggerated functional intensities characteristic of ALMFnet and CDDFuse. Zoomed-in regions further highlight FlexiD-Fuse's excellent tissue contrast and sharp edge preservation, notably superior to the overly smooth images of Diff-IF and SHIP. Additionally, FlexiD-Fuse adeptly merges high-frequency structural and low-frequency functional components, effectively delineating high-contrast PET regions alongside MRI anatomical structures. Lastly, FlexiD-Fuse demonstrates natural visual transitions, providing smoother multi-modal fusion compared to the abrupt transitions seen with DDcGAN and EMMA, thus enhancing clinical interpretability.

\add{\subsection{Computational Complexity Analysis}
As shown in Table \ref{tab:my-table}, compared with the current diffusion model-based methods (such as DDFM), the method proposed in this paper has fewer parameters and achieves superior visual fusion effects within a shorter inference time. Although there is still a certain gap in computational efficiency compared with some lightweight models based on encoders and decoders, this method demonstrates obvious advantages in high interpretability, flexible adaptability to modal inputs, and the visual quality of the fused images.}

\subsection{Extend experiment}
To rigorously verify the generalization fragility of the model beyond the medical imaging domain and the necessity of constructing a cross-domain evaluation system, this study systematically builds an extreme evaluation framework for heterogeneous physical properties by establishing three non-medical benchmark tests: infrared-visible light fusion, multi-focus fusion, and multi-exposure fusion. This experimental paradigm deliberately introduces cross-modal physical feature differences (between infrared and visible \add{images}) and the dynamic scalability challenge of input modalities (with multiple exposure images as input) to reveal whether the model trained solely in the medical field suffers from overfitting issues.  

\subsubsection{Infrared and Visible Image Fusion Experiment}
Infrared and visible image fusion improves image visibility, target recognition accuracy, and visual information completeness in complex environments by combining information from both modalities \cite{71,72,huang2024mma}.

In the infrared and visible fusion experiment, we comprehensively compared FlexiD-Fuse with current SOTA methods, including Diff-IF (Inform Fusion'24)\cite{36}, EMMA (CVPR'24) \cite{22}, Text-IF (CVPR'24)\cite{74}, SHIP (CVPR'24)\cite{69}, and CDDFuse (CVPR'23)\cite{20}, and evaluated them using three datasets: MSRS \cite{MSRS} (361 pairs), LLVIP \cite{LLVIP} (50 pairs), and M\textsuperscript{3}FD \cite{M3FD} (420 pairs).

\textbf{Objective Evaluation}. Table \ref{tab:4} shows the performance comparison of our method with other methods across the three datasets. Our model demonstrates significant superiority in the CC metric, indicating a high level of consistency between the fused results and source images. Additionally, the high scores in the FSIM metric further validate the model's advantage in preserving the naturalness of the images. Although not consistently leading in all metrics, overall, our method demonstrates strong performance in comprehensive evaluation, effectively fusing infrared and visible images to produce visually appealing results.
\begin{table}[h]
\centering
\caption{Objective metrics for different fusion methods on infrared and visible images (\textcolor{top1}{Red}: Best, \textcolor{top2}{Blue}: Second, and \textcolor{top3}{Green}: Third).}
\label{tab:4}
\resizebox{\columnwidth}{!}{%
\begin{tabular}{lccccc}
\toprule
\multicolumn{6}{c}{MSRS Dataset}                                         \\
\multicolumn{1}{c}{Methods} & VIF↑                                  & SSIM↑                                   & SCD↑                                   & FSIM↑                                  & CC↑                                    \\ \midrule
CDDFuse                     & 0.3993                               & \textcolor{top2}{0.7205}               & \textcolor{top3}{1.6442}             & 0.8240                               & \textcolor{top2}{0.6035}             \\
Diff-IF                     & \textcolor{top1}{0.4362}             & \textcolor{top1}{0.7252}               & 1.6242                               & \textcolor{top2}{0.8290}             & \textcolor{top3}{0.6023}             \\
EMMA                        & 0.3946                               & 0.7133                                & \textcolor{top1}{1.6530}             & 0.8222                               & 0.5997                               \\
SHIP                        & 0.4024                               & 0.6783                                & 1.5119                               & \textcolor{top3}{0.8293}             & 0.5945                               \\
Text-IF                     & \textcolor{top2}{0.4299}             & 0.7121                                & \textcolor{top2}{1.6460}             & \textcolor{top1}{0.8292}             & 0.5945                               \\
Our                         & \textcolor{top3}{0.4026}             & \textcolor{top3}{0.7157}               & 1.4504                               & 0.8270                               & \textcolor{top1}{0.6566}             \\ \midrule
\multicolumn{6}{c}{M\textsuperscript{3}FD Dataset}                                         \\
\multicolumn{1}{c}{Methods} & VIF↑                                  & SSIM↑                                   & SCD↑                                   & FSIM↑                                  & CC↑                                    \\ \midrule
CDDFuse                     & \textcolor{top2}{0.3838}                               & 0.5551                                & \textcolor{top1}{1.6956}             & 0.8202                               & \textcolor{top2}{0.5322}             \\
Diff-IF                     & 0.3524                               & \textcolor{top3}{0.7150}               & 1.3478                               & \textcolor{top2}{0.8222}                               & 0.4582                               \\
EMMA                        & 0.3371                               & 0.7074                                & \textcolor{top3}{1.5812}                               & 0.8173                               & \textcolor{top3}{0.5023}                               \\
SHIP                        & 0.3593                               & 0.7048                                & 1.3336                               & \textcolor{top1}{0.8236}             & 0.4516                               \\
Text-IF                     & \textcolor{top1}{0.3923}             & \textcolor{top2}{0.7216}               & 1.3966                               & 0.8213                               & 0.4598                               \\
Our                         & \textcolor{top3}{0.3828}             & \textcolor{top1}{0.7283}               & \textcolor{top2}{1.6930}             & \textcolor{top3}{0.8216}             & \textcolor{top1}{0.5682}             \\ \midrule
\multicolumn{6}{c}{LLVIP Dataset}                                        \\
\multicolumn{1}{c}{Methods} & VIF↑                                  & SSIM↑                                   & SCD↑                                   & FSIM↑                                  & CC↑                                    \\ \midrule
CDDFuse                     & \textcolor{top3}{0.3885}                               & 0.6817                                & \textcolor{top2}{1.6061}             & \textcolor{top3}{0.8217}                               & \textcolor{top2}{0.7603}             \\
Diff-IF                     & 0.3188                               & 0.6817                                & 1.3726                               & 0.8185                               & 0.7427                               \\
EMMA                        & 0.3642                               & \textcolor{top3}{0.7088}                                & \textcolor{top1}{1.6336}             & \textcolor{top2}{0.8240}             & \textcolor{top3}{0.7555}                               \\
SHIP                        & 0.3401                               & 0.6710                                & 1.3151                               & 0.8124                               & 0.7355                               \\
Text-IF                     & \textcolor{top2}{0.4029}             & \textcolor{top2}{0.7123}               & 1.3941                               & 0.8109                               & 0.7280                               \\
Our                         & \textcolor{top1}{0.4552}             & \textcolor{top1}{0.7219}               & \textcolor{top3}{1.4313}             & \textcolor{top1}{0.8306}             & \textcolor{top1}{0.7803}             \\ \bottomrule
\end{tabular}%
}
\end{table}
\begin{figure*}[h!]
\centering	
\includegraphics[width=1.0\linewidth]{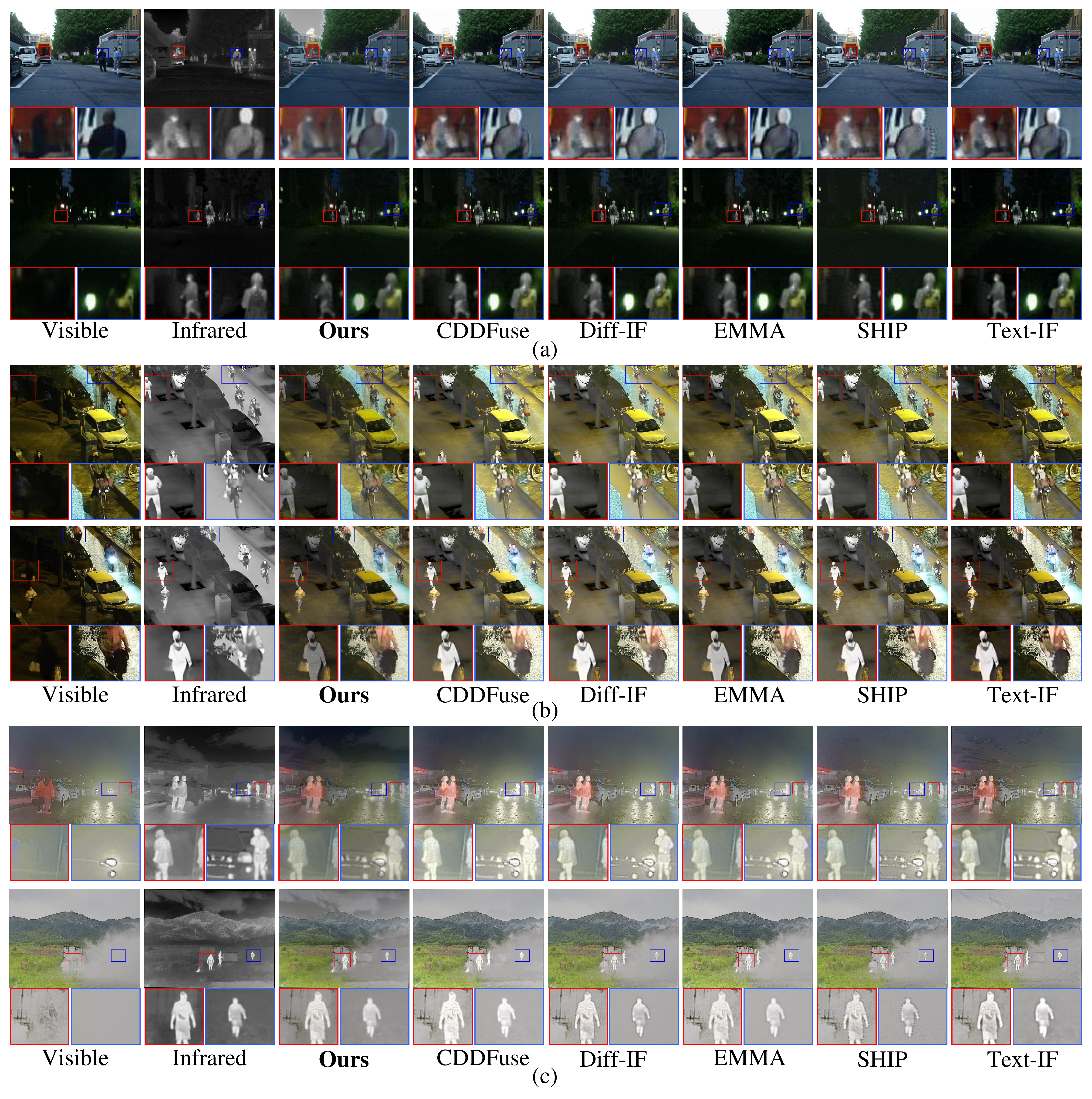}
\caption{Fusion results of different methods on infrared and visible image datasets, (a): MSRS, (b): LLVIP, (c): M\textsuperscript{3}FD.}
\label{Fig6}
\end{figure*}
\begin{table*}[ht]
\centering
\caption{Objective Metrics for \textbf{Multi-Exposure} of Different Methods on the SICE Dataset (\textcolor{top1}{Red}: Best, \textcolor{top2}{Blue}: Second, and \textcolor{top3}{Green}: Third)}
\label{tab:5}
\resizebox{\textwidth}{!}{%
\begin{tabular}{lccccccccc}
\toprule
Methods    & \(Q_{TE}\)↑                                 & \add{\(\mathrm{Q_{NCIE}}\)↑}                               & \(\mathrm{Q_{S}}\)↑                                 & VIF↑                                 & SSIM↑                                & PSNR↑                                & FSIM↑                                & SD↑                                  & CC↑                                 \\ \midrule
BHFMEF     & 0.5121& {0.8132}& \textcolor{top2}{0.7401}& 0.3369& \textcolor{top2}{0.5369}& 9.9449& 0.8088& 42.4976& 0.8530\\
HoloCo     & \textcolor{top3}{0.5278}& {0.8147}& 0.6504& {0.3755}& 0.5016& 10.011                           & \textcolor{top3}{0.8130}& 41.7803& \textcolor{top3}{0.8847}\\
HSDS-MEF   & 0.5180& \textcolor{top3}{0.8147}& \textcolor{top3}{0.7014}& \textcolor{top3}{0.3944}& 0.5248& \textcolor{top3}{10.1268}& 0.7927& \textcolor{top1}{50.4370}& {0.8824}\\
MERF       & 0.4962& 0.8108& 0.6813& 0.2891& 0.5055& 9.7865& 0.7755& 41.927                           & 0.8242\\
SAMF       & \textcolor{top1}{0.5493}& \textcolor{top1}{0.8198}& \textcolor{top1}{0.7459}& \textcolor{top1}{0.4878}& \textcolor{top1}{0.5561}& \textcolor{top1}{10.4359}& \textcolor{top1}{0.8240}& \textcolor{top3}{45.3329}& \textcolor{top1}{0.8939}\\
Ours        & \textcolor{top2}{0.5384}& \textcolor{top2}{0.8187}& 0.6858& \textcolor{top2}{0.4649}& \textcolor{top3}{0.5336}& \textcolor{top2}{10.2192}& \textcolor{top2}{0.8216}& \textcolor{top2}{48.0055}& \textcolor{top2}{0.8863}\\ \bottomrule
\end{tabular}%
}
\end{table*}

\begin{figure*}[h!]
\centering	
\includegraphics[width=1.0\linewidth]{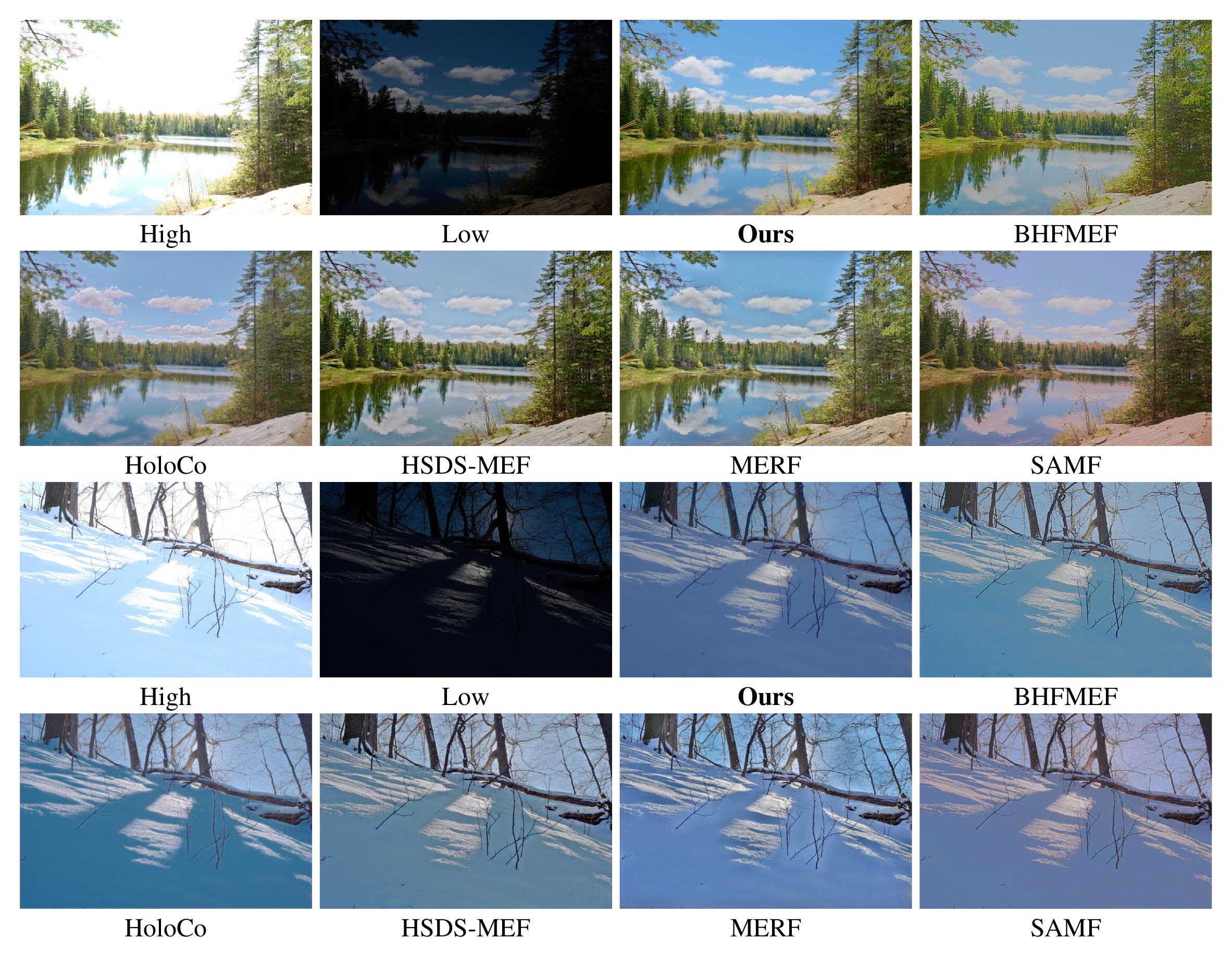}
\caption{Fusion results of different multi-exposure methods on the SICE dataset (Top: Spring, Bottom: Winter).}
\label{Fig7}
\end{figure*}

\textbf{Subjective Evaluation}. Figure \ref{Fig6} presents a qualitative comparison. Compared with existing methods, the proposed model effectively integrates visible light texture information and infrared thermal information, highlighting background details that are invisible under low-light conditions and clearly displaying the contours of people and vehicles. For person information obscured by fog, our method effectively captures human infrared features, achieving visual enhancement and providing a comprehensive understanding of the image content. SHIP and Diff-IF fail to effectively separate targets of interest from the background in complex scene fusion.

\begin{table*}[h!]
\centering
\caption{Objective Metrics for \textbf{Multi-Focus Fusion} Results of Different Methods on the MFI-WHU Dataset and Lytro. (\textcolor{top1}{Red}: Best, \textcolor{top2}{Blue}: Second, and \textcolor{top3}{Green}: Third)}
\label{tab:6,7}
\begin{tabular}{lccccc|lccccc}
\hline
\multirow{2}{*}{Methods} &
  \multicolumn{5}{c|}{\textbf{(a)}:Two Sources Fusion on MFI-WHU} &
  \multirow{2}{*}{Methods} &
  \multicolumn{5}{c}{\textbf{(b)}:Three Sources Fusion on Lytro} \\
 &
  {\add{\(\mathrm{Q_{NCIE}}\)↑}} &
  EN↑ &
  Nabf↑ &
  SD↑ &
  CC↑ &
   &
  {\add{\(\mathrm{Q_{NCIE}}\)↑}} &
  EN↑ &
  Nabf↑ &
  SD↑↑ &
  CC↑ \\ \hline
GRFuson &
  \textcolor{top1}{0.845} &
  \textcolor{top3}{7.322} &
  0.006 &
  52.37 &
  \textcolor{top1}{0.968} &
  GRFuson &
  \textcolor{top1}{0.829} &
  7.292 &
  0.041 &
  54.98 &
  \textcolor{top2}{0.975} \\
MUFusion &
  0.814 &
  \textcolor{top1}{7.509} &
  \textcolor{top1}{0.118} &
  \textcolor{top1}{63.76} &
  0.879 &
  MUFusion &
  0.816 &
  \textcolor{top1}{7.493} &
  \textcolor{top1}{0.145} &
  \textcolor{top1}{66.79} &
  0.894 \\
ZMFF &
  \textcolor{top2}{0.819} &
  7.268 &
  0.046 &
  \textcolor{top2}{58.09} &
  \textcolor{top3}{0.898} &
  ZMFF &
  \textcolor{top3}{0.822} &
  \textcolor{top3}{7.392} &
  \textcolor{top2}{0.093} &
  \textcolor{top2}{60.89} &
  0.935 \\
U2Fusion &
  \textcolor{top3}{0.816} &
  7.189 &
  0.024 &
  48.54 &
  \textcolor{top2}{0.953} &
  U2Fusion &
  0.820 &
  7.216 &
  0.020 &
  51.07 &
  \textcolor{top3}{0.974} \\
FusionDiff &
  0.813 &
  7.302 &
  \textcolor{top2}{0.090} &
  51.52 &
  0.736 &
  FusionDiff &
  \textcolor{top2}{0.824} &
  7.262 &
  0.043 &
  53.21 &
  \textcolor{top1}{0.979} \\
Ours &
  0.812 &
  \textcolor{top2}{7.336} &
  \textcolor{top3}{0.056} &
  \textcolor{top3}{52.56} &
  0.749 &
  Ours &
  0.814 &
  \textcolor{top2}{7.471} &
  \textcolor{top3}{0.058} &
  \textcolor{top3}{59.05} &
  0.870 \\ \hline
\end{tabular}
\end{table*}
\subsubsection{Multi-Exposure Experiment}
Multi-exposure image fusion overcomes the limitations of single images in dynamic range and detail representation by integrating information from different exposure images, generating more realistic and detailed images \cite{li2024single,jiang2025irw}. In this section, we compare the proposed method with several SOTA multi-exposure image fusion algorithms, including BHFMEF (ACM MM'23) \cite{75}, HoloCo (Information Fusion'23) \cite{76}, HSDS-MEF (AAAI'24) \cite{77}, MERF (ACM Trans.Graph'23) \cite{78}, and SAMF (ICASSP'24 Oral) \cite{79}.

\textbf{Objective evaluation}. Table \ref{tab:5} demonstrates the superior performance of our method across multiple evaluation metrics in multi-exposure image fusion. Although it does not achieve the highest scores in all metrics, the overall performance is balanced and robust, demonstrating the effectiveness and reliability of our method in fusion processing. The scores in VIF, SD, and PSNR indicate excellent performance in visual fidelity and multi-modal information integration, preserving rich feature details and sharp textures. Additionally, our model achieves very high scores in the CC metric, highlighting its outstanding performance in color consistency.

\textbf{Subjective Evaluation}. To enable better visual comparison, this subsection selects source images with distinct dominant colors, such as spring and winter scenery. As shown in Figure \ref{Fig7}, SAMF exhibits significant color deviation in global target fusion, with an overall yellowish tone, while HoloCo shows similar results and fails to restore vivid color saturation. HSDS-MEF introduces partial ghosting in the fusion results due to its dual-stream network's focus on feature fusion during intensity and illumination decomposition and recombination. Both MERF and our method excel in restoring image details and color richness, effectively handling distant image information and producing visually satisfying results.

\begin{figure*}[htbp]
\centering	
\includegraphics[width=1.0\linewidth]{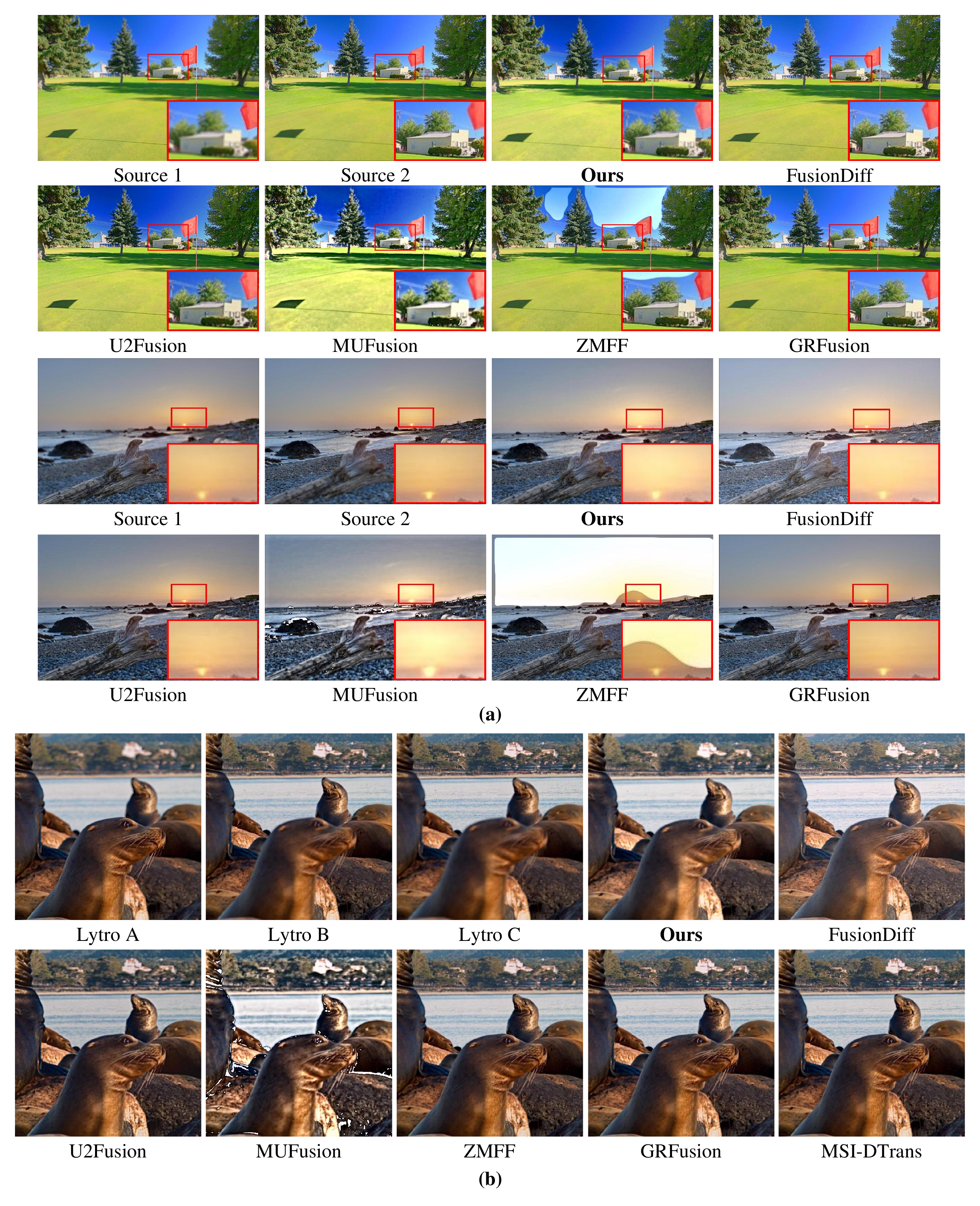}
\caption{(a) Fusion results of various methods on the MFI-WHU Dataset, with Source1 as the foreground-focused image and Source2 as the background-focused image. (b) Fusion results of various methods on the Lytro Three-Source Dataset, where Lytro A focuses on the foreground seal, Lytro B on the rear seal, and Lytro C on the background house.}
\label{Fig8}
\end{figure*}
\subsubsection{Multi-Focus Experiments}
Multi-focus image fusion aims to combine multiple images with different focal planes into a single clear image, ensuring that all target objects are clearly visible \cite{80,li2024focus,jiang2025refined,wang2025mmae}. In the multi-focus experiments, we performed two-source and multi-source image fusion on the MFI-WHU dataset \cite{MFI-WHU} and the Lytro dataset \cite{Lytro}. We compared our method with existing SOTA methods, including GRFusion (TCSVT'23) \cite{50}, MUFusion (Information Fusion'23) \cite{81}, U2Fusion (TPAMI'22) \cite{48}, FusionDiff (ESWA'24) \cite{82}, and ZMFF (Information Fusion'23) \cite{83}. Additionally, multi-source image fusion was further compared with MSI-DTrans (Displays'24) \cite{84}.

\begin{flushleft}
\textbf{Two Source Images Fusion}
\end{flushleft}
\textbf{Objective Evaluation}. As shown in Table \ref{tab:6,7} (a), our method achieves a good balance between detail preservation, contrast, and overall visual quality in multi-focus image fusion. Although slightly lower than other methods in certain metrics, it excels in entropy and standard deviation, indicating effective preservation of image details and contrast. Additionally, it achieves scores close to those of other high-performance methods in the \add{\(\mathrm{Q_{NCIE}}\)} and TMQI metrics, demonstrating good fusion quality.

\textbf{Subjective Evaluation}. In this section of the experiment, the fusion results are shown in Figure \ref{Fig8}(a). Observations reveal that the diffusion-based FlexiD-Fuse effectively preserves the focused information from the source images while avoiding the introduction of artifacts. It is noted that the MUFusion and FusionDiff methods exhibit color discrepancies, producing overall fused images that are brighter than the source images and somewhat overexposed. Additionally, MUFusion produces white spots in the fusion results, particularly around the trees behind the house. Due to a lack of complex background detection, ZMFF produces results with a significant amount of redundant background information, while U2Fusion and GRFusion deliver better visual performance.


\begin{table*}[h!]
\caption{Objective Metrics for Ablation Experiment Results (\textbf{Bold}: Best Performance)}
\label{Tab:8}
\centering
\begin{tabular}{ccccccccccc}
\hline
\multicolumn{3}{c|}{Configurations}                       & \multicolumn{8}{c}{Two-modal Medical Image Fusion}                                        \\ \cline{1-3}
  & Equation \ref{eq:15} & \multicolumn{1}{c|}{DFM Block} & VIF↑      & SSIM↑↑     & PSNR↑     & FSIM↑     & MI↑       & \(\mathrm{Q_{NCIE}}↑\) & \(\mathrm{Q_{S}}↑\) & \(\mathrm{Q_{CV}}\)↓ \\ \hline
1 & \ding{51}              & \ding{55}                        & 0.3754 & 0.6159 & 15.196 & 0.6183 & 0.7518 & 0.8019      & 0.8924    & 279.159    \\
2 &
  \ding{51} &
  \ding{51} &
  \textbf{0.3914} &
  \textbf{0.7493} &
  \textbf{16.839} &
  \textbf{0.7946} &
  \textbf{0.9209} &
  \textbf{0.8187} &
  \textbf{0.909} &
  \textbf{261.561} \\ \hline
\multicolumn{3}{c|}{Configurations}                       & \multicolumn{8}{c}{Tri-modal Medical Image Fusion}                                      \\ \cline{1-3}
  & Equation \ref{eq:15} & \multicolumn{1}{c|}{DFM Block} & VIF↑      & SSIM↑     & PSNR↑     & FSIM↑     & MI↑       & \add{\(\mathrm{Q_{NCIE}}\)↑} & \(\mathrm{Q_{S}}\)↑ & \(\mathrm{Q_{CV}}\)↓ \\ \hline
1 & \ding{55}              & \ding{51}                        & 7.1974 & 0.5648 & 45.196 & 0.8196 & 0.5978 & 0.8043       & 0.8942    & 1615.294   \\
2 & \ding{51}              & \ding{55}                        & 8.3498 & 0.6189 & 47.248 & 0.7891 & 0.4891 & 0.8012       & 0.8912    & 1689.159   \\
3 &
  \ding{51} &
  \ding{51} &
  \textbf{9.4243} &
  \textbf{0.6474} &
  \textbf{48.345} &
  \textbf{0.8909} &
  \textbf{0.6969} &
  \textbf{0.8191} &
  \textbf{0.8986} &
  \textbf{1596.948} \\ \hline
\end{tabular}
\end{table*}
\begin{flushleft}
\textbf{Multiple Source Images Fusion}
\end{flushleft}
\textbf{Objective evaluation}. In the three-image multi-focus fusion experiment, as shown in Table \ref{tab:6,7} (b), our method performs excellently in several key metrics, particularly entropy and standard deviation. This indicates that our method effectively preserves image detail and contrast, which is especially important for the clear presentation of multi-focus images. Although slightly lower than some methods in the CC and \add{\(\mathrm{Q_{NCIE}}\)} metrics, our method remains competitive in overall visual quality. Its higher entropy and moderate standard deviation ensure a good balance between information richness and visual quality.

\textbf{Subjective evaluation}. Unlike most existing multi-source fusion methods, the arbitrary modal fusion method proposed in this paper supports parallel processing of multiple source images for fusion. To evaluate the applicability of this method, this subsection not only compares it with the GRFusion method, which also supports multi-source image fusion, but also qualitatively compares it with methods focused on two-source image fusion. As shown in Figure \ref{Fig8} (b), the proposed method effectively achieves three-source image fusion, producing high-quality visual fusion results for details such as the dolphin's whiskers and background information. In contrast, the fusion results of GRFusion and MSI-DTrans exhibit over-enhancement and darker colors, while MUFusion fails to accurately handle focus information, with artifacts affecting the visual quality of the images.

\subsubsection{Summary of Extended Experiments}
In this section, the universality of the model is verified from multiple perspectives through non-medical experiments such as infrared and visible light fusion, multi-exposure image fusion, and multi-focus image fusion, enhancing its practicality in complex clinical scenarios. FlexiD-Fuse demonstrates outstanding performance in all experiments, being capable of efficiently fusing image information from different modalities and outperforming existing methods in terms of visual details, contrast, and consistency. The fusion results generated are more visually attractive. This further validates the flexible architecture and efficient reasoning ability of FlexiD-Fuse. Through sufficient verification in cross-domain experiments, it offers an expandable solution for multi-modal medical image fusion.

\subsection{Ablation Study}
The ablation experiments mainly focus on the effectiveness of the proposed DFM and EM modules in the innovation of the loss function, specifically addressing free-form input (Equation \ref{eq:15}). As shown in Table \ref{Tab:8}, the experimental results provide a comprehensive evaluation of 8 metrics for two-modal and tri-modal medical image fusion based on the Harvard dataset.

\textbf{DFM Block}. In this section, we validate the effectiveness of Mamba in FlexiD-Fuse. In Experiment 1, we replaced DFM with a CNN-based U-Net. To ensure fairness, both U-Net and DFM were trained with the same number of iterations and using the same dataset.

\textbf{Equation \ref{eq:15}}. This section validates the effectiveness of Equation \ref{eq:15}. In Experiment 2, \(img_3\) in Equation \ref{eq:15} was removed. For tri-modal fusion, a two-stage fusion approach was employed.

Overall, our method outperforms any experimental group, further demonstrating its effectiveness and superiority.

\section{Conclusion}
In this study, we propose FlexiD-Fuse, a novel diffusion-based framework designed for flexible multi-modal medical image fusion. Our method can flexibly handle the number of input images, effectively overcoming the limitation on input image numbers seen in existing techniques, \add{thus enabling clinicians to select the most suitable imaging modalities based on patient-specific diagnostic requirements, significantly enhancing clinical flexibility and diagnostic accuracy.} By integrating hierarchical Bayesian modeling and the Expectation-Maximization (EM) algorithm into the diffusion process, FlexiD-Fuse generates high-quality fused images without requiring ground-truth data. Extensive experiments conducted on bimodal, trimodal, and extended fusion tasks demonstrate the robustness and generalizability of our proposed framework.


\add{Despite the advantages of our study, we also acknowledge several limitations. First, compared to lightweight approaches, FlexiD-Fuse still incurs relatively high computational costs, which may hinder its deployment in resource-constrained environments such as embedded systems or real-time applications. Second, although the proposed method demonstrates strong generalization capabilities, its performance in multi-focus fusion scenarios still has room for improvement. Third, while FlexiD-Fuse currently supports flexible switching between two-modal and three-modal fusion, it is not yet capable of handling fusion tasks involving more than three modalities.

In future work, we plan to: (1) optimize the model architecture to reduce parameter count and computational complexity, thereby enabling efficient deployment on embedded devices and in real-time environments; (2) explore techniques such as mixture of experts (MoE) to further enhance the adaptability and overall performance of the framework; and (3) investigate the extension of our method to medical image fusion involving four or more modalities.}

\section*{Acknowledgement}
This research was supported by the National Natural Science Foundation of China (No. 62201149), the Natural Science Foundation of Guangdong Province (No. 2024A1515011880), the Basic and Applied Basic Research of Guangdong Province (No. 2023A1515140077), the Research Fund of Guangdong-HongKong-Macao Joint Laboratory for Intelligent Micro-Nano Optoelectronic Technology (No. 2020B1212030010).











\bibliographystyle{elsarticle-num} 
\bibliography{ref}



\end{document}